\documentclass[a4paper,10.5pt]{elsarticle}
\usepackage{geometry}
\usepackage{makecell}
\usepackage{extarrows}
\usepackage{subfigure}
\usepackage{mathrsfs}
\usepackage{lineno}
\usepackage{ulem}
\usepackage{graphicx}
\usepackage{subfigure}
\usepackage{makecell}
\usepackage{fancyhdr}
\usepackage{amssymb}
\usepackage{multicol}
\usepackage{amsmath}
\usepackage[figuresright]{rotating}

\usepackage{hyperref}
\usepackage{cleveref}

\usepackage{amsthm}
\theoremstyle{definition}

\newtheorem{thm}{Theorem}[section] 
\newtheorem{defn}[thm]{Definition} 
\newtheorem{pro}[thm]{Proposition} 
\newtheorem{exam}[thm]{Example}

\begin{document}
\begin{frontmatter}

\title{Foundational propositions of hesitant fuzzy soft $\beta$-covering approximation spaces}

\author{Shizhan Lu$^{1*}$}

\address{ \small  $^1$School of Management, Jiangsu University, Zhengjiang 212013, China\\ 


}


\begin{abstract}

Soft set theory serves as a mathematical framework for handling uncertain information, and hesitant fuzzy sets find extensive application in scenarios involving uncertainty and hesitation. 
Hesitant fuzzy sets exhibit diverse membership degrees, giving rise to various forms of inclusion relationships among them. 
This article introduces the notions of hesitant fuzzy soft $\beta$-coverings and hesitant fuzzy soft $\beta$-neighborhoods, which are formulated based on distinct forms of inclusion relationships among hesitancy fuzzy sets. Subsequently, several associated properties are investigated.
Additionally, specific variations of hesitant fuzzy soft $\beta$-coverings are introduced by incorporating hesitant fuzzy rough sets, followed by an exploration of properties pertaining to hesitant fuzzy soft $\beta$-covering approximation spaces. 

\end{abstract}
\end{frontmatter}

\begin{multicols}{2}
\section{Introduction}

The theory of Rough Set (RS) is highly significant when it comes to effective tools for handling uncertainty.
The RS theory, proposed by Pawlak \cite{PALA82}, extends the classical set theory.
RS focuses on lower and upper approximations and represents uncertain knowledge using known knowledge. The elementary structure of RS theory is an approximation space  that is composed of a given universe and an equivalence relation \cite{YUG}.
It has gained prominence as a burgeoning academic field in various domains such as machine learning \cite{KES, BJAT}, deep learning \cite{MTNA}, data mining \cite{CHLT, WGLT}, decision-making \cite{ZKZJ, JYE}, and others.

Subsequently, numerous variations of approximate spaces have been investigated, including fuzzy approximation spaces \cite{YUG}, fuzzy
$\beta$-covering approximation spaces \cite{ZXHWJQ, ZKDJ, ZHKDJ}, rough Pythagorean fuzzy approximate spaces \cite{MANH}, interval-valued fuzzy approximation spaces \cite{ZHLW, WJLX}, intuitionistic fuzzy approximation spaces \cite{HBGC, GSKGA}, intuitionistic fuzzy
$\beta$-covering approximation spaces \cite{HBLIH}, and interval-valued intuitionistic fuzzy approximation spaces \cite{MAMA}, among others.

The concept of hesitant fuzzy set, introduced by Torra  \cite{TV}, represents the membership degree of an element in a set as a multidimensional array.
Hesitant fuzzy sets concentrate on situations of uncertainty and hesitation, which are prevalent in real-world scenarios. 
Hesitant fuzzy sets find extensive applications in various domains, including decision-making \cite{HJH, LLYX, LNZQY, GXLY}, risk evaluation \cite{HXZS}, agglomerative hierarchical clustering \cite{JLGL}, investment policy \cite{DYM}, modeling for complex product manufacturing tasks \cite{BLYY}, linguistic perceptual \cite{AWNJE}, and forecasting \cite{HLJW, PMNME}, among others.

Nevertheless, a sufficiently clear definition of the inclusion relationship between two hesitant fuzzy sets is currently lacking. 
The definition of the inclusion relationship plays a crucial role in set theory since it establishes the equivalence of two sets when one is a subset of the other and vice versa. 
Babitha et al. \cite{BJ} proposed that $H_1$ is a hesitant fuzzy subset of $H_2$ if $H_1(x)\subset H_2(x)$ for all $x\in U$, but they did not provide an explicit interpretation of $H_1(x)\subset H_2(x)$. 
Carlos et al. \cite{JCVT} proposed a definition of the inclusion relationship that is overly specific and has limited applicability (only for the case $H_1^+(x)<H_2^-(x)$), failing to capture situations where both $H_1\subset H_2$ and $H_2\subset H_1$  hold simultaneously. 
Lu and Xu et al. \cite{LSZXZS} derived several types of inclusion relationships for hesitant fuzzy sets and presented foundational propositions based on the hesitant fuzzy membership degree in discrete form. 

Building upon the inclusion relationships and foundational propositions of hesitant fuzzy sets proposed by Lu and Xu et al. \cite{LSZXZS}, we investigate fundamental propositions related to hesitant fuzzy soft $\beta$-coverings, hesitant fuzzy soft $\beta$-neighborhoods, and hesitant fuzzy soft $\beta$-covering approximation spaces, among others.

The subsequent sections of this paper are organized as follows: Section 2 provides a comprehensive overview of the fundamental concepts of hesitant fuzzy sets. Section 3 presents several propositions concerning hesitant fuzzy soft $\beta$-neighborhoods and hesitant soft $\beta$-neighborhoods. Section 4 introduces various propositions related to hesitant fuzzy soft $\beta$-covering approximation spaces. Finally, Section 5 concludes the paper.

\section{Basic knowledge of hesitant fuzzy sets}

In this section, we briefly recall some basic concepts about hesitant fuzzy sets. If two sets $A$ and $B$ are classical sets, $A\sqcap B$ and $A\sqcup B$ represent the intersection and union of $A$ and $B$, respectively. Furthermore, $A\sqsubset B$ represents that $A$ is a subset of the classical set $B$.
Let $U$ be a universal set and $E$ be a set of parameters. Let $HF(U)$ be the set of all hesitant
fuzzy sets defined over $U$. 

\subsection{Reviews of hesitant fuzzy sets}

\begin{defn}\cite{TV} A hesitant fuzzy set in $U$ is defined as a function  that when applied to $U$ return a subset of $[0, 1]$.
\end{defn}

\begin{defn}\cite{TV} For each $x\in U$, and a hesitant fuzzy set $H$, we define the lower and upper bound
of $H(x)$ as

lower bound $H^-(x)=min\ H(x)$,

upper bound  $H^+(x)=max\ H(x)$.
\end{defn}

\begin{defn}\cite{TV} Given two hesitant fuzzy sets represented by their membership functions
$H_1$ and $H_2$, we define respectively their union and intersection as

union  $(H_1\cup H_2)(x)=\{h\in H_1(x)\sqcup H_2(x): h\geqslant max(H_1^-(x),H_2^-(x))\}$,

intersection $(H_1\cap H_2)(x)=\{h\in H_1(x)\sqcup H_2(x): h\leqslant min(H_1^+(x),H_2^+(x))\}$.
\end{defn}

\subsection{Inclusion relationships of hesitant fuzzy sets}

For two hesitant fuzzy sets $H_1$ and $H_2$, Babitha et al \cite{BJ} introduced that $H_1$ is a hesitant fuzzy subset of $H_2$ if $H_1(x)\subset H_2(x)$ for all $x\in U$. However, what is the meaning of $H_1(x)\subset H_2(x)$? There is no detailed description in \cite{BJ}.

Carlos et al \cite{JCVT} introduced a definition of inclusion relationship for hesitant fuzzy sets, i.e., $H_1\subset H_2$ $\Leftrightarrow$ $H_1\cup H_2=H_2$. By the description in \cite{JCVT}, $H_1(x)\subset H_2(x)\Leftrightarrow \{h:h\in H_2(x)\}=H_2(x)=(H_1\cup H_2)(x)=\{h\in H_1(x)\sqcup H_2(x):h\geqslant\max(H_1^-(x),H_2^-(x))\}$, i.e., $H_1(x)\subset H_2(x)$ implies $h<H_2^-(x)$ for all $h\in H_1(x)$. The definition of inclusion relationship for hesitant fuzzy sets in \cite{JCVT} is not perfect enough for these aspects, shown as (i) and (ii),

(i) The inclusion relationship of sets is also the ordering relationship of sets. A foundational and important proposition of inclusion relationship, $A\subset B$ and $B\subset A$ if and only if $A=B$, i.e., the equivalent sets contain each other. However, by the definition of inclusion relationship in \cite{JCVT}, if $H_1(x)\subset H_2(x)$ and $H_2(x)\subset H_1(x)$, then a contradiction is produced, i.e.,  $H_1^+(x)<H_2^-(x)<H_2^+(x)<H_1^-(x)<H_1^+(x)$.

(ii) The definition of inclusion relationship in \cite{JCVT} is too specific to have a small range of applications. For example, let $H_1(x)=\{0.1,0.2,0.5\}$, $H_2(x)=\{0.6,0.7,0.9\}$ and $H_3(x)=\{0.1,0.5,0.8\}$, then $H_1(x)\subset H_2(x)$ is obvious. However, this definition cannot describe the relationship of $H_2(x)$ and $H_3(x)$ and is not applicable for many cases of hesitant fuzzy sets.

Lu and Xu et al. \cite{LSZXZS} derived several types of inclusion relationships for hesitant fuzzy sets and presented foundational propositions based on the hesitant fuzzy membership degree in discrete form. The descriptions in \cite{LSZXZS} are shown as Example \ref{examstart} and Definition \ref{def2}.

\begin{exam}\label{examstart}\cite{LSZXZS} Let $U=\{x_1,x_2,x_3,x_4,x_5,x_6\}$ be a set of decision-making schemes and $H$ be an expert team that consists of three experts. $H(U)=\frac{0.9,0.2}{x_1}+\frac{0.6,0.6,0.5}{x_2}+\frac{0.7,0.5,0.5}{x_3}+\frac{0.8,0.6,0.5}{x_4}+\frac{0.9,0.3,0.1}{x_5}+\frac{0.9,0.8,0.7}{x_6}$ are the  schemes' scores that are given by experts, in which the scores for $x_1$ are 0.9 and 0.2 that are obtained through the evaluations made by experts. One of the three experts fails to evaluate the scheme $x_1$.

(1) Here, $x_1$ has a score of 0.9, which is greater than or equal to all the scores for $x_2$, it is possible that the scheme $x_1$ is better than the scheme $x_2$, denoted as $H(x_2)\subset_pH(x_1)$.

(2) Here, $0.55=mean[H(x_1)]<mean[H(x_2)]=0.567$, where $mean[\cdot]$ is the mean value operator. In comparing the mean values of scores, we find that the scheme $x_2$ is better than the scheme $x_1$, denoted as $H(x_1)\subset_mH(x_2)$.

(3) On the one hand, the best score of $x_3$ is 0.7, which is greater than or equal to the best score of $x_2$. On the other hand, the worst score of $x_3$ is 0.5, which
is greater than or equal to the worst score of $x_2$. To compare the respective best and worst cases of schemes $x_2$ and $x_3$, it is acceptable that the scheme $x_3$ is better than the scheme $x_2$, denoted as $H(x_2)\subset_aH(x_3)$.

(4) To compare the scores for schemes $x_3$ and $x_4$ one by one ($0.7\leqslant 0.8$; $0.5\leqslant 0.6$; and $0.5\leqslant0.5$), it is strongly credible that the scheme $x_4$ is better than the scheme $x_3$, denoted as $H(x_3)\subset_sH(x_4)$.

(5) We can obtain $H(x_1)\subset_sH(x_5)$ after truncating the tail score of $x_5$, i.e., deleting the score 0.1 of $x_5$, this case is denoted as $H(x_1)\subset_{st}H(x_5)$ and is recorded briefly as $H(x_1)\subset_{t}H(x_5)$.

(6) The worst score of $x_6$ is greater than or equal to the best score of $x_3$. Thus, it is necessary
that the scheme $x_6$ is better than the scheme $x_3$, which is denoted as $H(x_3)\subset_nH(x_6)$.

Comparing the mean value of scores is a common thought for decision-making;  however, while doing so, some important information may be lost, such as the best and the worst scores.
\end{exam}

\begin{defn}\label{def2}\cite{LSZXZS} Let $H_1$ and $H_2$ be two hesitant fuzzy sets on $U$. Several kinds of inclusion relationships of two hesitant fuzzy sets are defined as follows,

(1) If $H^+_1(x)\leqslant H^+_2(x)$, then $H_1(x)\subset_p H_2(x)$. If $H_1(x)\subset_p H_2(x)$ for all $x\in U$, then $H_1\subset_p H_2$. If $H_1\subset_p H_2$ and $H_2\subset_p H_1$, then $H_1=_p H_2$.

(2) If $H^+_1(x)\leqslant H^+_2(x)$ and $H^-_1(x)\leqslant H^-_2(x)$, then $H_1(x)\subset_a H_2(x)$. If $H_1(x)\subset_a H_2(x)$ for all $x\in U$, then $H_1\subset_a H_2$. If $H_1\subset_a H_2$ and $H_2\subset_a H_1$, then $H_1=_a H_2$.

(3) If $mean[H_1(x)]\leqslant mean[H_2(x)]$, then $H_1(x)\subset_m H_2(x)$. If $H_1(x)\subset_m H_2(x)$ for all $x\in U$, then $H_1\subset_m H_2$. If $H_1\subset_m H_2$ and $H_2\subset_m H_1$, then $H_1=_m H_2$.

(4) Let $H_1(x)=V=\{v_1,v_2,\cdots,v_k\}$  and $H_2(x)=W=\{w_1,w_2,\cdots,w_l\}$ be two descending sequences. If $k\geqslant l$ and $w_i\geqslant v_i$ for $1\leqslant i\leqslant l$, then $H_1(x)\subset_s H_2(x)$. If $H_1(x)\subset_s H_2(x)$ for all $x\in U$, then $H_1\subset_s H_2$. If $H_1\subset_s H_2$ and $H_2\subset_s H_1$, then $H_1=_s H_2$.

(5) Let $H_1(x)=V=\{v_1,v_2,\cdots,v_k\}$  and $H_2(x)=W=\{w_1,w_2,\cdots,w_l\}$ be two descending sequences. If  $k<l$ and $w_i\geqslant v_i$ for $1\leqslant i\leqslant k$, then $H_1(x)\subset_{t}H_2(x)$. If $H_1(x)\subset_{t} H_2(x)$ for all $x\in U$, then $H_1\subset_{t} H_2$. It is obvious that $H_1\subset_{t} H_2$ and $H_2\subset_{t} H_1$ cannot hold simultaneously.

(6) If $H_1^+(x)\leqslant H_2^-(x)$, then $H_1(x)\subset_n H_2(x)$. If $H_1(x)\subset_n H_2(x)$ for all $x\in U$, then $H_1\subset_n H_2$.  If $H_1\subset_n H_2$ and $H_2\subset_n H_1$, then $H_1=_n H_2$.

\end{defn}

\subsection{Foundational propositions of hesitant fuzzy sets}

\begin{pro}\label{pro2.7}\cite{LSZXZS} Suppose $A,B,C\in HF(U)$, the following statements hold,

(1) If $A\subset_p B$ and $B\subset_p C$, then $A\subset_p C$.

(2) If $A\subset_a B$ and $B\subset_a C$, then $A\subset_a C$.

(3) If $A\subset_m B$ and $B\subset_m C$, then $A\subset_m C$.

(4) If $A\subset_s B$ and $B\subset_s C$, then $A\subset_s C$.

(5) If $A\subset_t B$ and $B\subset_t C$, then $A\subset_t C$.

(6) If $A\subset_n B$ and $B\subset_n C$, then $A\subset_n C$.

\end{pro}

\begin{pro}\label{pro9ch}\cite{LSZXZS} The following statements hold for $A,B\in HF(U)$,

(1) If $A\subset_a B$, then $A\subset_p B$.

(2) If $A\subset_s B$, then $A\subset_a B$.

(3) If $A\subset_s B$, then $A\subset_m B$.

(4) If $A\subset_t B$, then $A\subset_p B$.

(5) If  $A\subset_n B$, then one of $A(x)\subset_s B(x)$ and $A(x)\subset_t B(x)$ holds for all $x\in U$.

\end{pro}

\begin{pro}\label{pro2.17}\cite{LSZXZS} Suppose $A,B,C\in HF(U)$, the following statements hold,

(1) $A\subset_p B$ and $A\subset_p C$ if and only if $A\subset_p B\cap C$.

(2) $A\subset_a B$ and $A\subset_a C$ if and only if $A\subset_a B\cap C$.

(3) $A\subset_t B$ and $A\subset_t C$, then  $A\subset_t B\cap C$.

(4) $A\subset_n B$ and $A\subset_n C$ if and only if $A\subset_n B\cap C$.
\end{pro}

\begin{defn} \cite{TV} Given a hesitant fuzzy set represented by its membership function $h$ we
define its complement as follows
\end{defn}

$h^c(x)=\bigcup\limits_{\gamma\in h(x)}\{1-\gamma\}$.

\begin{pro}\cite{LSZXZS} Suppose $A,B\in HF(U)$, the following statements hold,

(1) $A\subset_a B$, then $B^c\subset_a A^c$.

(2) $A\subset_m B$, then $B^c\subset_m A^c$.

(3) $A\subset_s B$, then $B^c\subset_{sot} A^c$.

(4) $A\subset_n B$, then $B^c\subset_n A^c$.
\end{pro}

\subsection{Equivalence relationships of hesitant fuzzy sets}

\begin{thm}\label{thm1}\cite{LSZXZS}  Let $A,B\in HF(U)$ and $x\in U$, the following statements hold,

(1) $A= B$, then $A=_p B$, $A=_a B$ and $A=_m B$.

(2) $A=_s B$ $\Leftrightarrow$ $A= B$.

(3) $A=_s B$, then $A=_p B$, $A=_a B$ and $A=_m B$.

(4) If $A=_n B$, then $h'=h''$ for all $h'\in A(x)$ and all $h''\in B(x)$.

(5) If $A=_n B$, then $A=_p B$, $A=_a B$ and $A=_m B$.
\end{thm}

\begin{thm}\label{thm6}\cite{LSZXZS}  Suppose $A,B,C\in HF(U)$, the following statements hold,

(1) $A\cap A=_p A$, $A\cup A=_p A$.

(2) $A\cap A=_a A$, $A\cup A=_a A$.

(3) $A\cap A=_m A$, $A\cup A=_m A$.

(4) $(A\cup B)\cap A=_p A$, $(A\cap B)\cup A=_p A$.

(5) $(A\cup B)\cap A=_a A$, $(A\cap B)\cup A=_a A$.

(6) $(A\cup B)\cap C=_p (C\cap A)\cup(C\cap B)$, $(A\cap B)\cup C=_p (C\cup A)\cap(C\cup B)$.

(7) $(A\cup B)\cap C=_a (C\cap A)\cup(C\cap B)$, $(A\cap B)\cup C=_a (C\cup A)\cap(C\cup B)$.
\end{thm}

\begin{thm}\label{thm8} \cite{TV,XX,BJ} Suppose $A,B,C\in HF(U)$, the follows are held.

(1) $A\cap B= B\cap A$, $A\cup B= B\cup A$.

(2) $(A\cap B)\cap C= A\cap(B\cap C)$, $(A\cup B)\cup C= A\cup(B\cup C)$.

(3) $(A^c)^c= A$.

(4) $(A\cap B)^c= A^c\cup B^c$.

(5) $(A\cup B)^c= A^c\cap B^c$.
\end{thm}

\subsection{Families of hesitant fuzzy sets}

Suppose $\mathscr{H}_1$ and $\mathscr{H}_2$ are two families of hesitant fuzzy sets. If $H_i\in \mathscr{H}_2$ holds for all $H_i\in \mathscr{H}_1$, we say that $\mathscr{H}_1$ is the classical subset of $\mathscr{H}_2$, denoted as $\mathscr{H}_1\sqsubset \mathscr{H}_2$.

\begin{thm}\label{thm2.26}\cite{LSZXZS}  Let $A\in HF(U)$ and $\mathscr{H}\sqsubset HF(U)$ be a family of  hesitant fuzzy sets over $U$. The following statements hold,

(1) $A\subset_p H$ for all $H\in\mathscr{H}$ if and only if $A\subset_p\bigcap\{H: H\in \mathscr{H}\}$.

(2) If $A\subset_p H_{\alpha}$ for a $H_{\alpha}\in\mathscr{H}$, then $A\subset_p\bigcup\{H: H\in \mathscr{H}\}$.

(3) $A\subset_a H$ for all $H\in\mathscr{H}$ if and only if $A\subset_a\bigcap\{H: H\in \mathscr{H}\}$.

(4) If $A\subset_a H_{\alpha}$ for a $H_{\alpha}\in\mathscr{H}$, then $A\subset_a\bigcup\{H: H\in \mathscr{H}\}$.

(5) If $A\subset_t H$ for all $H\in\mathscr{H}$, then $A\subset_t\bigcap\{H: H\in \mathscr{H}\}$.

(6) If $A\subset_t H_{\alpha}$ for a $H_{\alpha}\in\mathscr{H}$, and $|A(x)|<|H(x)|$ for all $H\in\mathscr{H}$ and $x\in U$, then $A\subset_t\bigcup\{H: H\in \mathscr{H}\}$.

(7) $A\subset_n H$ for all $H\in\mathscr{H}$ if and only if $A\subset_n\bigcap\{H: H\in \mathscr{H}\}$.

(8) If $A\subset_n H_{\alpha}$ for a $H_{\alpha}\in\mathscr{H}$, then $A\subset_n\bigcup\{H: H\in \mathscr{H}\}$.
\end{thm}

\begin{thm}\label{thm3}\cite{LSZXZS} Let $\mathscr{H}_1,\mathscr{H}_2\sqsubset HF(U)$ be two families of  hesitant fuzzy sets over $U$. The following statements hold,

(1) If $\mathscr{H}_1\sqsubset\mathscr{H}_2$, then $\bigcap\{H: H\in \mathscr{H}_2\}\subset_p\bigcap\{H: H\in \mathscr{H}_1\}$.

(2) If $\mathscr{H}_1\sqsubset\mathscr{H}_2$, then $\bigcup\{H: H\in \mathscr{H}_1\}\subset_p\bigcup\{H: H\in \mathscr{H}_2\}$.

(3)  If $\mathscr{H}_1\sqsubset\mathscr{H}_2$, then $\bigcap\{H: H\in \mathscr{H}_1\}\subset_p\bigcup\{H: H\in \mathscr{H}_2\}$.

(4) If $\mathscr{H}_1\sqsubset\mathscr{H}_2$, then $\bigcap\{H: H\in \mathscr{H}_2\}\subset_a\bigcap\{H: H\in \mathscr{H}_1\}$.

(5) If $\mathscr{H}_1\sqsubset\mathscr{H}_2$, then $\bigcup\{H: H\in \mathscr{H}_1\}\subset_a\bigcup\{H: H\in \mathscr{H}_2\}$.

(6)  If $\mathscr{H}_1\sqsubset\mathscr{H}_2$, then $\bigcap\{H: H\in \mathscr{H}_1\}\subset_a\bigcup\{H: H\in \mathscr{H}_2\}$.

(7) If $\mathscr{H}_1\sqsubset\mathscr{H}_2$, then $\bigcup\{H: H\in \mathscr{H}_1\}\subset_{sot}\bigcup\{H: H\in \mathscr{H}_2\}$.

(8)  If $\mathscr{H}_1\sqsubset\mathscr{H}_2$, then $\bigcap\{H: H\in \mathscr{H}_1\}\subset_{sot}\bigcup\{H: H\in \mathscr{H}_2\}$.
\end{thm}

\section{Hesitant fuzzy soft sets and hesitant fuzzy soft $\beta$-neighborhood}

\subsection{Hesitant fuzzy soft sets}

\begin{defn} \cite{BJ}  Let $U$ be an initial universe set and $E$ be a set of parameters. A pair $(F,E)$ is a hesitant fuzzy soft sets if $F(e)\in HF(U)$  for every $e\in E$.
\end{defn}

\begin{defn}   For two hesitant fuzzy soft sets $(F,A)$ and $(G,B)$ over a common universe $U$,
$A\sqsubset B\sqsubset E$, for every $e\in A$,

(1) $(F,A)\subset_p (G,B)$ if $F(e)\subset_p G(e)$.

(2) $(F,A)\subset_a (G,B)$ if $F(e)\subset_a G(e)$.

(3) $(F,A)\subset_m (G,B)$ if $F(e)\subset_m G(e)$.

(4) $(F,A)\subset_s (G,B)$ if $F(e)\subset_s G(e)$.

(5) $(F,A)\subset_t (G,B)$ if $F(e)\subset_t G(e)$.

(6) $(F,A)\subset_n (G,B)$ if $F(e)\subset_n G(e)$.
\end{defn}

\begin{defn}  Let $A,B,C\sqsubseteq E$, $(F,A)$ and $(G,B)$ be two hesitant fuzzy soft sets over $U$.

(1)  The intersection of $(F,A)$ and $(G,B)$, is denoted by $(H,C)=(F,A)\cap(G,B)$, where $C=A\sqcap B$ and $H(e)=F(e)\cap G(e)$ for every $e\in C$.

(2)  The union of $(F,A)$ and $(G,B)$, is denoted by $(H,C)=(F,A)\widetilde{\cup}(G,B)$,  where $C=A\sqcup B$ and
$$H(e)= \left\{\begin{array}{lll}
   F(e),  \ \ \ \ \ \ \ \ \ \ \ \  e\in A-B, \\
   F(e)\cup G(e), \ \ \ e\in A\sqcap B,\\
   G(e),\ \ \  \ \ \ \ \ \ \ \ \ e\in B-A.
\end{array}
\right.$$

 (3) The restricted union of $(F,A)$ and $(G,B)$, is denoted by $(H,C)=(F,A)\cup(G,B)$, where $C=A\sqcap B$ and $H(e)=F(e)\cup G(e)$ for every $e\in C$.
\end{defn}

The restricted intersection of $(F,A)$ and $(G,B)$ is defined as (1), and used the same mark $(H,C)=(F,A)\cap(G,B)$.

\begin{defn}  \cite{BJ}  Let $(F,A)$ be a hesitant fuzzy soft set. The compliment of $(F,A)$ is denoted by $(F,A)^c$, and $(F,A)^c=(F^c,A)$, where $F^c(a)$ is the compliment of the hesitant fuzzy set $F(a)$.
\end{defn}

It is obvious that $((F,A)^c)^c=(F,A)$.

\begin{exam} Let $U=\{x,y\}$, $E=\{e_1,e_2,e_3,e_4,\newline e_5\}$, $(F,E)$ and $(G,E)$ are shown in Table 1.
\end{exam}

\begin{tabular}{p{38pt}|p{50pt} p{50pt}}
\multicolumn{3}{c}{Table 1. $(F,E)$ and $(G,E)$.}\\
\hline
$(F,E)$  &  $x$  &   $y$  \\
\hline
$e_1$ & 0.2,0.3,0.5 & 0.3,0.5,0.7  \\
$e_2$ & 0.2,0.3,0.4 & 0.1,0.2,0.3  \\
$e_3$ & 0.5,0.6,0.8 & 0.1,0.5,0.8  \\
$e_4$ & 0.3,0.5     & 0.2,0.6      \\
$e_5$ & 0.3,0.5,0.6 & 0.1,0.2,0.3  \\
\hline
$(G,E)$  &  $x$  &   $y$  \\
\hline
$e_1$ & 0.1,0.2,0.6 & 0.2,0.5,0.8  \\
$e_2$  & 0.2,0.4,0.5 & 0.2,0.3,0.3  \\
$e_3$  & 0.6,0.6,0.9 & 0.2,0.5,0.8  \\
$e_4$  & 0.5,0.6,0.7 & 0.7,0.8,0.9  \\
$e_5$ & 0.7,0.8,0.9 & 0.3,0.4,0.5  \\
\hline
\end{tabular}
\vspace{3mm}

Let $A_1=\{e_1,e_2\}$, $A_2=\{e_1,e_2,e_3\}$, then $(F,A_1)\subset_p(G,A_2)$.

Let $A_3=\{e_2,e_3\}$, $A_4=\{e_2,e_3,e_4\}$, then $(F,A_3)\subset_a(G,A_4)$, $(F,A_3)\subset_m(G,A_4)$ and $(F,A_3)\subset_s(G,A_4)$.

Let $A_5=\{e_4\}$, $A_6=\{e_4,e_5\}$, then $(F,A_5)\subset_t(G,A_6)$.

Let $A_7=\{e_4,e_5\}$, $A_8=\{e_3,e_4,e_5\}$, then $(F,A_7)\subset_n(G,A_8)$.

Let $A=\{e_1,e_2\}$, $B=\{e_2,e_3\}$, $\mathscr{H}_1$ be the family of hesitant fuzzy soft sets of $(F,A)\cap(G,B)$, $\mathscr{H}_2$ be the family of hesitant fuzzy soft sets of $(F,A)\cup(G,B)$, $\mathscr{H}_3$ be the family of hesitant fuzzy soft sets of  $(F,A)\widetilde{\cup}(G,B)$.
$$\mathscr{H}_1=\begin{cases}
H_{11}(e_2)=\frac{0.2,0.2,0.3,0.4,0.4}{x}+\frac{0.1,0.2,0.2,0.3,0.3,0.3}{y}.
\end{cases}$$
$$\mathscr{H}_2=\begin{cases}
H_{21}(e_2)=\frac{0.2,0.2,0.3,0.4,0.4,0.5}{x}+\frac{0.2,0.2,0.3,0.3,0.3}{y}.
\end{cases}$$
$$\mathscr{H}_3=\begin{cases}
H_{31}(e_1)=\frac{0.2,0.3,0.5}{x}+\frac{0.3,0.5,0.7}{y},\\
H_{32}(e_2)=\frac{0.2,0.2,0.3,0.4,0.4,0.5}{x}+\frac{0.2,0.2,0.3,0.3,0.3}{y},\\
H_{33}(e_3)=\frac{0.6,0.6,0.9}{x}+\frac{0.2,0.5,0.8}{y}.\\
\end{cases}$$

The attribute domains of definition of $\mathscr{H}_1$ and $\mathscr{H}_2$ are same, i.e., $\{e_2\}$. The attribute domains of definition of  $\mathscr{H}_3$ is $\{e_1,e_2,e_3\}$.  The attribute domains of definition of $\mathscr{H}_1$ and $\mathscr{H}_3$ are not same. $\mathscr{H}_2$ is a classical subset of $\mathscr{H}_3$, $\mathscr{H}_2\sqsubseteq \mathscr{H}_3$.

\begin{thm}\label{thm10}  \cite{BJ}  Let $(F,A)$ and $(G,B)$ be  two  hesitant fuzzy soft sets over $U$. Then

(1) $[(F,A)\cup(G,B)]^c=(F,A)^c\cap(G,B)^c$,

(2) $[(F,A)\cap(G,B)]^c=(F,A)^c\cup(G,B)^c$.
\end{thm}

It should be noted that the Theorem \ref{thm10} is only applicative for the restricted union and restricted intersection of two hesitant fuzzy soft sets. The domains of definition of the union and intersection of two hesitant fuzzy soft sets are not same, the Theorem \ref{thm10} is not applicative for the union and intersection of two hesitant fuzzy soft sets.

\begin{defn}  Let $A,B\sqsubseteq E$, $(F,A)$ and $(G,B)$ be two hesitant fuzzy soft sets over $U$. Let
$$\begin{cases}
(F,A)\wedge(G,B)=F(e_i)\cap G(e_j),\ e_i\in A,\ e_j\in B,\\
(F,A)\vee(G,B)=F(e_i)\cup G(e_j),\ e_i\in A,\ e_j\in B.\\
\end{cases}$$
\end{defn}

\begin{thm}\label{thm11}  Let $(F,A)$ and $(G,B)$ be  two  hesitant fuzzy soft sets over $U$. Then

(1) $[(F,A)\vee(G,B)]^c=(F,A)^c\wedge(G,B)^c$,

(2) $[(F,A)\wedge(G,B)]^c=(F,A)^c\vee(G,B)^c$.
\end{thm}

{\bf\slshape Proof} For all $e_i\in A$ and $e_j\in B$, $F(e_i)$ and $G(e_j)$ are two hesitant fuzzy sets. By Theorem \ref{thm8}, $[F(e_i)\cup G(e_j)]^c=F(e_i)^c\cap G(e_j)^c$ and $[F(e_i)\cap G(e_j)]^c=F(e_i)^c\cup G(e_j)^c$, then Theorem \ref{thm11} holds.$\blacksquare$

\begin{thm}\label{thm12}  Let $A,B,C\sqsubseteq E$, $(F,A)$, $(G,B)$ and $(H,C)$ be three hesitant fuzzy soft sets over $U$. Then

(1) $(F,A)\cap (F,A)=_p (F,A)$,

(2) $(F,A)\cup (F,A)=_p (F,A)$.

(3) $(F,A)\cap (F,A)=_a (F,A)$,

(4) $(F,A)\cup (F,A)=_a (F,A)$.

(5) $(F,A)\cap (F,A)=_m (F,A)$,

(6) $(F,A)\cup (F,A)=_m (F,A)$.

(7) $((F,A)\cup (G,B))\cap (F,A)=_p (F,A)$,

(8) $((F,A)\cup (G,B))\cap (F,A)=_a (F,A)$.

(9) $((F,A)\cap (G,B))\cup (F,A)=_p (F,A)$,

(10) $((F,A)\cap (G,B))\cup (F,A)=_a (F,A)$.

(11) $((F,A)\cup (G,B))\cap (H,C)=_p ((H,C)\cap (F,A))\cup((H,C)\cap (G,B))$,

(12) $((F,A)\cup (G,B))\cap (H,C)=_a ((H,C)\cap (F,A))\cup((H,C)\cap (G,B))$.

(13) $((F,A)\cap (G,B))\cup (H,C)=_p ((H,C)\cup (F,A))\cap((H,C)\cup (G,B))$,

(14) $((F,A)\cap (G,B))\cup (H,C)=_a ((H,C)\cup (F,A))\cap((H,C)\cup (G,B))$.

(15) $(F,A)\cap (G,B)= (G,B)\cap (F,A)$,

(16) $(F,A)\cup (G,B)= (G,B)\cup (F,A)$.

(17) $((F,A)\cap (G,B))\cap (H,C)= (F,A)\cap((G,B)\cap (H,C))$,

(18) $((F,A)\cup (G,B))\cup (H,C)= (F,A)\cup((G,B)\cup (H,C))$.
\end{thm}

{\bf\slshape Proof} For all $e_i\in A$, $e_j\in B$ and $e_k\in C$, $F(e_i)$, $G(e_j)$ and $H(e_k)$ are three hesitant fuzzy sets. By Theorem \ref{thm6} and \ref{thm8}, Theorem \ref{thm12} holds.$\blacksquare$

\subsection{Hesitant fuzzy soft $\beta$-neighborhood}

Let the hesitant fuzzy number $\beta$ be an indicator to survey other hesitant fuzzy sets in the processes of decision-making. If $H^-(x)=1$ for all $x\in U$, then $H$ is denoted as $H^U$. If $H^+(x)=1$ for all $x\in U$, then $H$ is denoted as $H^{pU}$.  If $H^+(x)=0$ for all $x\in U$, then $H$ is denoted as $H^{\emptyset}$. If $A^-(x^*)=1$ for a $x^*\in U$, then $A(x^*)$ is denoted as $\mathbf{1}(x^*)$.

\begin{center}
\begin{table*}
\centering
\label{table}
\setlength{\tabcolsep}{3pt}
\begin{tabular}{l|l|l|l|l|llll}
\multicolumn{6}{c}{Table 2. A triple $(U,F,E)$.}\\
\hline
$U$ &$x_1$ &$x_2$ &$x_3$    & $x_4$ &$x_5$   \\
\hline
$e_1$ & 0.5, 0.4, 0.3\  & 1, 1                &1, 1, 1       &1, 1, 0.2          & 0.7, 0.3, 0.2          \\
$e_2$ & 1, 1                & 0.4, 0.3, 0.2       &0.5, 0.3, 0.3          & 1, 1, 1           &  1, 1, 1     \\
$e_3$ &   0.7, 0.5, 0.2           &  0.5, 0.4 & 0.5        & 0.5, 0.4  &      0.6, 0.5, 0.2          \\
$e_4$ &  0.8, 0.7           &    0.6, 0.1        & 0.9, 0.8, 0.2          & 0.6, 0.5           &  0.6, 0.6      \\
$e_5$ & 0.8, 0.7, 0.7       & 0.6, 0.5, 0.4, 0.3              & 0.9, 0.8   & 0.7, 0.6, 0.5, 0.1     & 0.8, 0.7, 0.6, 0.1  \\
$e_6$ & 0.6, 0.6       & 0.7, 0.5        & 0.7, 0.6  & 0.8, 0.7        & 0.7, 0.6     \\
$e_7$ &0.4, 0.3       & 0.3, 0.2             & 0.2, 0.1  & 0.4, 0.3         & 0.3, 0.2      \\
$e_8$ &0.7, 0.7, 0.5, 0.4        & 0.6, 0.5, 0.4, 0.4            &  0.8, 0.6, 0.5, 0.5  &0.7, 0.7, 0.5, 0.4    & 0.7, 0.6, 0.5, 0.5  \\
$e_9$ &   0.7, 0.5, 0.2           &  0.5, 0.4 & 0.5        & 0.5, 0.4  &      0.6, 0.5, 0.3          \\
\hline
\end{tabular}
\label{tab1}
\end{table*}
\end{center}

\begin{defn} Suppose $U$ and $E$ are the sets of objects and parameters, respectively. Let $A\sqsubset E$, $F:E\rightarrow HF(U)$, $F(e)\in HF(U)$,  $e\in A$. If $\mathbf{1}\subset_{sot}\bigcup\limits_{e\in A}F(e)$, then we called $(F,A)$ a hesitant fuzzy soft covering over $U$, where $\mathbf{1}\in HF(U)$, $\mathbf{1}^-(x)=1$ and $|\mathbf{1}(x)|=\min\{|F(e)(x)|:e\in A\}$.
\end{defn}

Here, $(U,F,A)$ is called a hesitant fuzzy soft covering approximation space.

\begin{defn} Let $\beta$ be a hesitant fuzzy number, $F:E\rightarrow HF(U)$, $e\in A\sqsubset E$.

(1) If $\beta\subset_p(\bigcup\limits_{e\in A}F(e))(x)$ for all $x\in U$, then $(F,A)_p$ is called a hesitant fuzzy soft $p\beta$-covering over $U$, the triple $T_p=(U,F,A)_{\beta}$ is called a hesitant fuzzy soft $p\beta$-covering approximation space.

(2) If $\beta\subset_a(\bigcup\limits_{e\in A}F(e))(x)$ for all $x\in U$, then $(F,A)_a$ is called a hesitant fuzzy soft $a\beta$-covering over $U$, the triple $T_a=(U,F,A)_{\beta}$ is called a hesitant fuzzy soft $a\beta$-covering approximation space.

(3) If $\beta\subset_m(\bigcup\limits_{e\in A}F(e))(x)$ for all $x\in U$, then $(F,A)_m$ is called a hesitant fuzzy soft $m\beta$-covering over $U$, the triple $T_m=(U,F,A)_{\beta}$ is called a hesitant fuzzy soft $m\beta$-covering approximation space.

(4) If $\beta\subset_s(\bigcup\limits_{e\in A}F(e))(x)$ for all $x\in U$, then $(F,A)_s$ is called a hesitant fuzzy soft $s\beta$-covering over $U$, the triple $T_s=(U,F,A)_{\beta}$ is called a hesitant fuzzy soft $s\beta$-covering approximation space.

(5) If $\beta\subset_t(\bigcup\limits_{e\in A}F(e))(x)$ for all $x\in U$, then $(F,A)_t$ is called a hesitant fuzzy soft $t\beta$-covering over $U$, the triple $T_t=(U,F,A)_{\beta}$ is called a hesitant fuzzy soft $t\beta$-covering approximation space.

(6) If $\beta\subset_n(\bigcup\limits_{e\in A}F(e))(x)$ for all $x\in U$, then $(F,A)_n$ is called a hesitant fuzzy soft $n\beta$-covering over $U$, the triple $T_n=(U,F,A)_{\beta}$ is called a hesitant fuzzy soft $n\beta$-covering approximation space.
\end{defn}

\begin{defn}\label{def5} Let $\beta$ be a hesitant fuzzy number, and $U$ and $E$ be the sets of objects and parameters, respectively. Let $e\in A\sqsubset E$, $F:E\rightarrow HF(U)$.

(1) $\widetilde{SN}^{p\beta}_x=\bigcap\{F(e)\in F(A):\beta\subset_p F(e)(x)\}$ is called a hesitant fuzzy soft $p\beta$-neighborhood of $x$ for $A\in T_p=(U,F,A)_{\beta}$.

(2) $\widetilde{SN}^{a\beta}_x=\bigcap\{F(e)\in F(A):\beta\subset_a F(e)(x)\}$ is called a hesitant fuzzy soft $a\beta$-neighborhood of $x$ for $A\in T_a=(U,F,A)_{\beta}$.

(3)  $\widetilde{SN}^{m\beta}_x=\bigcap\{F(e)\in F(A):\beta\subset_m F(e)(x)\}$ is called a hesitant fuzzy soft $m\beta$-neighborhood of $x$ for $A\in T_m=(U,F,A)_{\beta}$.

(4)  $\widetilde{SN}^{s\beta}_x=\bigcap\{F(e)\in F(A):\beta\subset_s F(e)(x)\}$ is called a hesitant fuzzy soft $s\beta$-neighborhood of $x$ for   $A\in T_s=(U,F,A)_{\beta}$.

(5)  $\widetilde{SN}^{t\beta}_x=\bigcap\{F(e)\in F(A):\beta\subset_t F(e)(x)\}$ is called a hesitant fuzzy soft $t\beta$-neighborhood of $x$ for   $A\in T_t=(U,F,A)_{\beta}$.

(6) $\widetilde{SN}^{n\beta}_x=\bigcap\{F(e)\in F(A):\beta\subset_n F(e)(x)\}$ is called a hesitant fuzzy soft $n\beta$-neighborhood of $x$ for  $A\in T_n=(U,F,A)_{\beta}$.

\end{defn}

\begin{exam}\label{examp1} The triple $(U,F,E)$ are shown in Table 2. Let $\beta=\{0.5,0.4,0.3\}$.

(1) Let $A=\{e_1,e_2\}$. $(U,F,A)$ is a hesitant fuzzy soft covering approximation space.

$|\mathbf{1}(x_1)|=\min\{|F(e)(x_1)|:e\in A\}=\min\{3,2\}=2$, $|\mathbf{1}(x_2)|=2$, $|\mathbf{1}(x_3)|=3$, $|\mathbf{1}(x_4)|=3$, $|\mathbf{1}(x_5)|=3$. Then $\mathbf{1}=\frac{1,1}{x_1}+\frac{1,1}{x_2}+\frac{1,1,1}{x_3}+\frac{1,1,1}{x_4}+\frac{1,1,1}{x_5}$. $\bigcup\limits_{e\in A}F(e)=\frac{1,1}{x_1}+\frac{1,1}{x_2}+\frac{1,1,1}{x_3}+\frac{1,1,1,1,1}{x_4}+\frac{1,1,1}{x_5}$. $\mathbf{1}\subset_{sot}\bigcup\limits_{e\in A}F(e)$.

(2) Let $A=\{e_1,e_2\}$. Since $(\bigcup\limits_{e\in A}F(e))^+(x_i)=1$ for all $x_i\in U$, then $\beta\subset_p(\bigcup\limits_{e\in A}F(e))(x_i)$. $T_p=(U,F,A)_{\beta}$ is a hesitant fuzzy soft $p\beta$-covering approximation space.

In this $T_p=(U,F,A)_{\beta}$, $A=\{e_1,e_2\}$, $\beta\subset_p F(e_1)(x_1)$ and $\beta\subset_p F(e_2)(x_1)$, then $\widetilde{SN}^{p\beta}_{x_1}=F(e_1)\cap F(e_2)=\frac{0.5,0.4,0.3}{x_1}+\frac{0.4,0.3,0.2}{x_2}+\frac{0.5,0.3,0.3}{x_3}+\frac{1,1,1,1,1,0.2}{x_4}+\frac{0.7,0.3,0.2}{x_5}$.

$\beta\subset_p F(e_1)(x_2)$ and $\beta\not\subset_p F(e_2)(x_2)$, then $\widetilde{SN}^{p\beta}_{x_2}=F(e_1)=\frac{0.5,0.4,0.3}{x_1}+\frac{1,1}{x_2}+\frac{1,1,1}{x_3}+\frac{1,1,0.2}{x_4}+\frac{0.7,0.3,0.2}{x_5}$.

(3) Let $A=\{e_1,e_2\}$. $\beta\subset_a(\bigcup\limits_{e\in A}F(e))(x_i)$ for all $x_i\in U$, then $T_a=(U,F,A)_{\beta}$ is a hesitant fuzzy soft $a\beta$-covering approximation space.

In this $T_a$, $\beta\subset_a F(e_1)(x_1)$ and $\beta\subset_a F(e_2)(x_1)$, then $\widetilde{SN}^{a\beta}_{x_1}=F(e_1)\cap F(e_2)$. $\beta\subset_a F(e_1)(x_2)$ and $\beta\not\subset_a F(e_2)(x_2)$, then $\widetilde{SN}^{a\beta}_{x_2}=F(e_1)$.

(4) Let $A=\{e_2,e_3\}$. $\bigcup\limits_{e\in A}F(e)=\frac{1,1}{x_1}+\frac{0.5,0.4,0.4}{x_2}+\frac{0.5,0.5}{x_3}+\frac{1,1,1}{x_4}+\frac{1,1,1}{x_5}$. $\beta\subset_m(\bigcup\limits_{e\in A}F(e))(x)$ for all $x\in U$, then $T_m=(U,F,A)_{\beta}$ is a hesitant fuzzy soft $m\beta$-covering approximation space.

In this $T_m$, $\beta\subset_m F(e_2)(x_1)$ and $\beta\subset_m F(e_3)(x_1)$, then $\widetilde{SN}^{m\beta}_{x_1}=F(e_2)\cap F(e_3)$. $\beta\not\subset_m F(e_2)(x_2)$ and $\beta\subset_m F(e_3)(x_2)$, then $\widetilde{SN}^{m\beta}_{x_2}=F(e_3)$.

(5) Let $A=\{e_3,e_4\}$. $\bigcup\limits_{e\in A}F(e)=\frac{0.8,0.7,0.7}{x_1}+\frac{0.6,0.5,0.4}{x_2}+\frac{0.9,0.8,0.5}{x_3}+\frac{0.6,0.5,0.5}{x_4}+\frac{0.6,0.6,0.6}{x_5}$. $\beta\subset_s(\bigcup\limits_{e\in A}F(e))(x)$ for all $x\in U$, then $T_s=(U,F,A)_{\beta}$ is a hesitant fuzzy soft $s\beta$-covering approximation space.

In this $T_s$, $\beta\not\subset_s F(e_3)(x_1)$ and $\beta\subset_s F(e_4)(x_1)$, then $\widetilde{SN}^{s\beta}_{x_1}=F(e_4)$. $\beta\subset_s F(e_3)(x_2)$ and $\beta\not\subset_s F(e_4)(x_2)$, then $\widetilde{SN}^{s\beta}_{x_2}=F(e_3)$.

(6) Let $A=\{e_4,e_5\}$. $\bigcup\limits_{e\in A}F(e)=\frac{0.8,0.8,0.7,0.7,0.7}{x_1}+\frac{0.6,0.6,0.5,0.4,0.3}{x_2}+\frac{0.9,0.9,0.8,0.8}{x_3}+\frac{0.7,0.6,0.6,0.5,0.5}{x_4}+\frac{0.8,0.7,0.6,0.6,0.6}{x_5}$. $\beta\subset_t(\bigcup\limits_{e\in A}F(e))(x)$ for all $x\in U$, then $T_t=(U,F,A)_{\beta}$ is a hesitant fuzzy soft $t\beta$-covering approximation space.

In this $T_t$, $\beta\not\subset_t F(e_4)(x_1)$ and $\beta\not\subset_t F(e_5)(x_1)$, then $\widetilde{SN}^{t\beta}_{x_1}=Null$ (i.e., $\widetilde{SN}^{t\beta}_{x_1}$ is undefined in this $T_t$). $\beta\not\subset_t F(e_4)(x_2)$ and $\beta\subset_t F(e_5)(x_2)$, then $\widetilde{SN}^{t\beta}_{x_2}=F(e_5)$.

(7) Let $A=\{e_5,e_6\}$. $\bigcup\limits_{e\in A}F(e)=\frac{0.8,0.7,0.7}{x_1}+\frac{0.7,0.6,0.5,0.5}{x_2}+\frac{0.9,0.8}{x_3}+\frac{0.8,0.7,0.7}{x_4}+\frac{0.8,0.7,0.7,0.6,0.6}{x_5}$. $\beta\subset_n(\bigcup\limits_{e\in A}F(e))(x)$ for all $x\in U$, then $T_n=(U,F,A)_{\beta}$ is a hesitant fuzzy soft $n\beta$-covering approximation space.

In this $T_n$, $\beta\subset_n F(e_5)(x_1)$ and $\beta\subset_n F(e_6)(x_1)$, then $\widetilde{SN}^{n\beta}_{x_1}=F(e_5)\cap F(e_6)$. $\beta\not\subset_n F(e_5)(x_5)$ and $\beta\subset_n F(e_6)(x_5)$, then $\widetilde{SN}^{n\beta}_{x_5}=F(e_6)$.
\end{exam}

\begin{pro}\label{pro7} Let $\beta$ be a hesitant fuzzy number, the following statements hold for all $x\in U$.

(1) For each $T_p=(U,F,A)_{\beta}$, $\beta\subset_p\widetilde{SN}^{p\beta}_x(x)$.

(2) For each $T_a=(U,F,A)_{\beta}$, $\beta\subset_a\widetilde{SN}^{a\beta}_x(x)$.

(3) For each $T_t=(U,F,A)_{\beta}$, $\beta\subset_t\widetilde{SN}^{t\beta}_x(x)$.

(4) For each $T_n=(U,F,A)_{\beta}$, $\beta\subset_n\widetilde{SN}^{n\beta}_x(x)$.
\end{pro}

{\bf\slshape Proof} (1) For each $x\in U$, $\widetilde{SN}^{p\beta}_x=\bigcap\{F(e)\in F(A):\beta\subset_p F(e)(x)\}$. By Theorem \ref{thm2.26} (1), if $\beta\subset_p F(e)(x)$ for all $e\in A^*\sqsubset A$, then $\beta\subset_p (\bigcap\limits_{e\in A^*}F(e))(x)$, i.e., $\beta\subset_p\widetilde{SN}^{p\beta}_x(x)$.

(2) By Theorem \ref{thm2.26} (3), it can be proved similarly to (1).

(3) By Theorem \ref{thm2.26} (5), it can be proved similarly to (1).

(4) By Theorem \ref{thm2.26} (7), it can be proved similarly to (1).$\blacksquare$

\begin{exam} Let $U=\{x\}$, $\beta=\{0.5,0.4,0.3,0.2,\newline 0.1\}$, $F(e_1)=\frac{0.5,0.4,0.3}{x}$, $F(e_2)=\frac{0.6,0.6,0.6,0.6,0.1}{x}$, $F(e_3)=\frac{0.6,0.6,0.6,0.6,0.1,0.1}{x}$, $F(e_4)=\frac{0.5,0.1}{x}$, $F(e_5)=\frac{0.3,0.3}{x}$.

$\beta\subset_m F(e_4)(x)$ and $\beta\subset_m F(e_5)(x)$, however $\beta\not\subset_m (F(e_4)\cap F(e_5))(x)$.

$\beta\subset_s F(e_1)(x)$ and $\beta\subset_s F(e_2)(x)$, however $\beta\not\subset_s (F(e_1)\cap F(e_2))(x)$.

$\beta\subset_s F(e_1)(x)$ and $\beta\subset_t F(e_3)(x)$, however $\beta\not\subset_{sot} (F(e_1)\cap F(e_3))(x)$.
\end{exam}

\begin{pro}\label{prop1} Let $\beta$ be a hesitant fuzzy number,  for all $x,y,z\in U$.

(1) For each $T_p$, if $\beta\subset_p\widetilde{SN}^{p\beta}_x(y)$ and $\beta\subset_p\widetilde{SN}^{p\beta}_y(z)$, then $\beta\subset_p\widetilde{SN}^{p\beta}_x(z)$.

(2) For each $T_a$, if $\beta\subset_a\widetilde{SN}^{a\beta}_x(y)$ and $\beta\subset_a\widetilde{SN}^{a\beta}_y(z)$, then $\beta\subset_a\widetilde{SN}^{a\beta}_x(z)$.

(3) For each $T_n$, if $\beta\subset_n\widetilde{SN}^{n\beta}_x(y)$ and $\beta\subset_n\widetilde{SN}^{n\beta}_y(z)$, then $\beta\subset_n\widetilde{SN}^{n\beta}_x(z)$.
\end{pro}

{\bf\slshape Proof} (1) For each $e\in A\in T_p$, $\beta\subset_p\widetilde{SN}^{p\beta}_x(y)=(\bigcap\limits_{\beta\subset_p F(e)(x)}F(e))(y)$.
By the necessity of Theorem \ref{thm2.26} (1), we can obtain that $\beta\subset_p F(e)(y)$ holds for each $F(e)\in\{F(e):\beta\subset_p F(e)(x)\}$. It means that if $\beta\subset_p F(e)(x)$ holds then $\beta\subset_p F(e)(y)$ holds for each $e\in A$.

In the similar way, $\beta\subset_p\widetilde{SN}^{p\beta}_y(z)$ implies that if $\beta\subset_p F(e)(y)$ holds then $\beta\subset_p F(e)(z)$ holds for each $e\in A$.

To sum up,  if $\beta\subset_p F(e)(x)$ holds then $\beta\subset_p F(e)(z)$ holds  for each $e\in A$.
By the sufficiency of Theorem \ref{thm2.26} (1), $\beta\subset_p(\bigcap\{F(e):\beta\subset_p F(e)(x)\})(z)=\widetilde{SN}^{p\beta}_x(z)$.

(2) By Theorem \ref{thm2.26} (3),  it can be proved similarly to (1).

(3) By Theorem \ref{thm2.26} (7),  it can be proved similarly to (1).$\blacksquare$

\begin{pro}\label{prop2} For two hesitant fuzzy number $\beta_1$ and $\beta_2$, the following statements hold for all $x\in U$.

(1) For each $T_p$, if $\beta_1\subset_p\beta_2$, then $\widetilde{SN}^{p\beta_1}_x\subset_a\widetilde{SN}^{p\beta_2}_x$.

(2)  For each $T_a$, if $\beta_1\subset_a\beta_2$, then $\widetilde{SN}^{a\beta_1}_x\subset_a\widetilde{SN}^{a\beta_2}_x$.

(3)  For each $T_m$, if $\beta_1\subset_m\beta_2$, then $\widetilde{SN}^{m\beta_1}_x\subset_a\widetilde{SN}^{m\beta_2}_x$.

(4) For each $T_s$, if $\beta_1\subset_s\beta_2$, then $\widetilde{SN}^{s\beta_1}_x\subset_a\widetilde{SN}^{s\beta_2}_x$.

(5) For each $T_t$, if $\beta_1\subset_t\beta_2$, then $\widetilde{SN}^{t\beta_1}_x\subset_a\widetilde{SN}^{t\beta_2}_x$.

(6) For each $T_n$, if $\beta_1\subset_n\beta_2$, then $\widetilde{SN}^{n\beta_1}_x\subset_a\widetilde{SN}^{n\beta_2}_x$.
\end{pro}

{\bf\slshape Proof} (1) Since $\beta_1\subset_p \beta_2$,  if $\beta_2\subset_p F(e)(x)$ for $e\in A\in T_p$, by Proposition \ref{pro2.7} (1),  $\beta_1(x)\subset_p F(e)(x)$ holds, i.e., $\{F(e):\beta_2\subset_p F(e)(x),e\in A\}\sqsubset\{F(e):\beta_1\subset_p F(e)(x),e\in A\}$.

By Theorem \ref{thm3} (4), $\{F(e):\beta_2\subset_p F(e)(x),e\in A\}\sqsubset\{F(e):\beta_1\subset_p F(e)(x),e\in A\}$ implies $\widetilde{SN}^{p\beta_1}_x=\bigcap\{F(e):\beta_1\subset_p F(e)(x),e\in A\}\subset_a\bigcap\{F(e):\beta_2\subset_p F(e)(x),e\in A\}=\widetilde{SN}^{p\beta_2}_x$.

(2) Since $\beta_1\subset_a \beta_2$,  if $\beta_2\subset_a F(e)(x)$ for $e\in A\in T_p$, by Proposition \ref{pro2.7} (2),  $\beta_1(x)\subset_a F(e)(x)$ holds, i.e., $\{F(e):\beta_2\subset_a F(e)(x),e\in A\}\sqsubset\{F(e):\beta_1\subset_a F(e)(x),e\in A\}$. By Theorem \ref{thm3} (4), $\widetilde{SN}^{a\beta_1}_x=\bigcap\{F(e):\beta_1\subset_a F(e)(x),e\in A\}\subset_a\bigcap\{F(e):\beta_2\subset_a F(e)(x),e\in A\}=\widetilde{SN}^{a\beta_2}_x$.

(3) Since $\beta_1\subset_m \beta_2$,  if $\beta_2\subset_m F(e)(x)$ for $e\in A\in T_m$, by Proposition \ref{pro2.7} (3),  $\beta_1(x)\subset_m F(e)(x)$ holds, i.e., $\{F(e):\beta_2\subset_m F(e)(x),e\in A\}\sqsubset\{F(e):\beta_1\subset_m F(e)(x),e\in A\}$. By Theorem \ref{thm3} (4), $\widetilde{SN}^{m\beta_1}_x=\bigcap\{F(e):\beta_1\subset_m F(e)(x),e\in A\}\subset_a\bigcap\{F(e):\beta_2\subset_m F(e)(x),e\in A\}=\widetilde{SN}^{m\beta_2}_x$.

(4) Since $\beta_1\subset_s \beta_2$,  if $\beta_2\subset_s F(e)(x)$ for $e\in A\in T_s$, by Proposition \ref{pro2.7} (4),  $\beta_1(x)\subset_s F(e)(x)$ holds, i.e., $\{F(e):\beta_2\subset_s F(e)(x),e\in A\}\sqsubset\{F(e):\beta_1\subset_s F(e)(x),e\in A\}$. By Theorem \ref{thm3} (4), $\widetilde{SN}^{s\beta_1}_x=\bigcap\{F(e):\beta_1\subset_s F(e)(x),e\in A\}\subset_a\bigcap\{F(e):\beta_2\subset_s F(e)(x),e\in A\}=\widetilde{SN}^{s\beta_2}_x$.

(5)  Since $\beta_1\subset_t \beta_2$,  if $\beta_2\subset_t F(e)(x)$ for $e\in A\in T_t$, by Proposition \ref{pro2.7} (5),  $\beta_1(x)\subset_t F(e)(x)$ holds, i.e., $\{F(e):\beta_2\subset_t F(e)(x),e\in A\}\sqsubset\{F(e):\beta_1\subset_t F(e)(x),e\in A\}$. By Theorem \ref{thm3} (4), $\widetilde{SN}^{t\beta_1}_x=\bigcap\{F(e):\beta_1\subset_t F(e)(x),e\in A\}\subset_a\bigcap\{F(e):\beta_2\subset_t F(e)(x),e\in A\}=\widetilde{SN}^{t\beta_2}_x$.

(6)  Since $\beta_1\subset_n \beta_2$,  if $\beta_2\subset_n F(e)(x)$ for $e\in A\in T_n$, by Proposition \ref{pro2.7} (6),  $\beta_1(x)\subset_n F(e)(x)$ holds, i.e., $\{F(e):\beta_2\subset_n F(e)(x),e\in A\}\sqsubset\{F(e):\beta_1\subset_n F(e)(x),e\in A\}$. By Theorem \ref{thm3} (4), $\widetilde{SN}^{n\beta_1}_x=\bigcap\{F(e):\beta_1\subset_n F(e)(x),e\in A\}\subset_a\bigcap\{F(e):\beta_2\subset_n F(e)(x),e\in A\}=\widetilde{SN}^{n\beta_2}_x$.$\blacksquare$

\begin{pro}\label{pro8} Let $\beta$ be a hesitant fuzzy number,  $x,y\in U$.

(1) For each $T_p$, $\beta\subset_p \widetilde{SN}^{p\beta}_x(y)$ if and only if $\widetilde{SN}^{p\beta}_y\subset_p \widetilde{SN}^{p\beta}_x$.

(2) For each $T_p$, $\beta\subset_p \widetilde{SN}^{p\beta}_x(y)$ and $\beta\subset_p \widetilde{SN}^{p\beta}_y(x)$ if and only if $\widetilde{SN}^{p\beta}_x=_p\widetilde{SN}^{p\beta}_y$.

(3) For each $T_a$, $\beta\subset_a \widetilde{SN}^{a\beta}_x(y)$ if and only if $\widetilde{SN}^{a\beta}_y\subset_a \widetilde{SN}^{a\beta}_x$.

(4) For each $T_a$, $\beta\subset_a \widetilde{SN}^{a\beta}_x(y)$ and $\beta\subset_a \widetilde{SN}^{a\beta}_y(x)$ if and only if $\widetilde{SN}^{a\beta}_x=_a\widetilde{SN}^{a\beta}_y$.

(5) For each $T_n$, $\beta\subset_n \widetilde{SN}^{n\beta}_x(y)$, then $\widetilde{SN}^{n\beta}_y\subset_a \widetilde{SN}^{n\beta}_x$.

(6) For each $T_n$, $\beta\subset_n \widetilde{SN}^{n\beta}_x(y)$ and $\beta\subset_n \widetilde{SN}^{n\beta}_y(x)$, then  $\widetilde{SN}^{n\beta}_x=_a\widetilde{SN}^{n\beta}_y$.
\end{pro}

{\bf\slshape Proof} (1) For $e\in A\in T_p$, if $\beta\subset_p \widetilde{SN}^{p\beta}_x(y)=(\bigcap\limits_{\beta\subset_p F(e)(x)}F(e))(y)$, by the necessity of Theorem \ref{thm2.26} (1), $\beta\subset_p F(e)(y)$ holds for each $F(e)\in\{F(e):\beta\subset_p F(e)(x)\}$, then,  $\{F(e):\beta\subset_p F(e)(x)\}\sqsubset\{F(e):\beta\subset_p F(e)(y)\}$ is obtained.

By Theorem \ref{thm3} (1), $\{F(e):\beta\subset_p F(e)(x)\}\sqsubset\{F(e):\beta\subset_p F(e)(y)\}$ implies $\bigcap\{F(e):\beta\subset_p F(e)(y)\}\subset_p \bigcap\{F(e):\beta\subset_p F(e)(x)\}$, i.e., $\widetilde{SN}^{p\beta}_y\subset_p \widetilde{SN}^{p\beta}_x$.

On the other hand, by Proposition \ref{pro7} (1), $\beta\subset_p\widetilde{SN}^{p\beta}_y(y)$ holds. If $\widetilde{SN}^{p\beta}_y\subset_p \widetilde{SN}^{p\beta}_x$, i.e., $\widetilde{SN}^{p\beta}_y(y)\subset_p \widetilde{SN}^{p\beta}_x(y)$, by Proposition \ref{pro2.7} (1), $\beta\subset_p\widetilde{SN}^{p\beta}_y(y)\subset_p\widetilde{SN}^{p\beta}_x(y)$.

(2) By the results of (1), $\beta\subset_p \widetilde{SN}^{p\beta}_x(y)$ and $\beta\subset_p \widetilde{SN}^{p\beta}_y(x)$ if and only if $\widetilde{SN}^{p\beta}_y\subset_p \widetilde{SN}^{p\beta}_x$ and $\widetilde{SN}^{p\beta}_x\subset_p \widetilde{SN}^{p\beta}_y$, i.e., $\widetilde{SN}^{p\beta}_y=_p \widetilde{SN}^{p\beta}_x$.

(3) In the similar way of (1), if $\beta\subset_a \widetilde{SN}^{a\beta}_x(y)=(\bigcap\limits_{\beta\subset_a F(e)(x)}F(e))(y)$, by the necessity of Theorem \ref{thm2.26} (3),  $\{F(e):\beta\subset_a F(e)(x)\}\sqsubset\{F(e):\beta\subset_a F(e)(y)\}$ holds. By Theorem \ref{thm3} (4), $\bigcap\{F(e):\beta\subset_a F(e)(y)\}\subset_a \bigcap\{F(e):\beta\subset_a F(e)(x)\}$, i.e., $\widetilde{SN}^{a\beta}_y\subset_a \widetilde{SN}^{a\beta}_x$.

On the other hand, by Proposition \ref{pro7} (2), $\beta\subset_a\widetilde{SN}^{a\beta}_y(y)$ holds. If $\widetilde{SN}^{a\beta}_y\subset_a \widetilde{SN}^{a\beta}_x$, i.e., $\widetilde{SN}^{a\beta}_y(y)\subset_a \widetilde{SN}^{a\beta}_x(y)$, by Proposition \ref{pro2.7} (2), $\beta\subset_a\widetilde{SN}^{a\beta}_y(y)\subset_a\widetilde{SN}^{a\beta}_x(y)$.

(4) By the results of (3), $\beta\subset_a \widetilde{SN}^{a\beta}_x(y)$ and $\beta\subset_a \widetilde{SN}^{a\beta}_y(x)$ if and only if $\widetilde{SN}^{a\beta}_y\subset_a \widetilde{SN}^{a\beta}_x$ and $\widetilde{SN}^{a\beta}_x\subset_a \widetilde{SN}^{a\beta}_y$, i.e., $\widetilde{SN}^{a\beta}_y=_a \widetilde{SN}^{a\beta}_x$.

(5) In the similar way of (1), if $\beta\subset_n \widetilde{SN}^{n\beta}_x(y)=(\bigcap\limits_{\beta\subset_n F(e)(x)}F(e))(y)$, by the necessity of Theorem \ref{thm2.26} (7),  $\{F(e):\beta\subset_n F(e)(x)\}\sqsubset\{F(e):\beta\subset_n F(e)(y)\}$ holds. By Theorem \ref{thm3} (4), $\bigcap\{F(e):\beta\subset_n F(e)(y)\}\subset_a \bigcap\{F(e):\beta\subset_n F(e)(x)\}$,  i.e., $\widetilde{SN}^{n\beta}_y\subset_a \widetilde{SN}^{n\beta}_x$.

(6) By the results of (5), $\beta\subset_n \widetilde{SN}^{n\beta}_x(y)$ and $\beta\subset_n \widetilde{SN}^{n\beta}_y(x)$ imply $\widetilde{SN}^{n\beta}_y\subset_a\widetilde{SN}^{n\beta}_x$ and $\widetilde{SN}^{n\beta}_x\subset_a\widetilde{SN}^{n\beta}_y$, i.e., $\widetilde{SN}^{n\beta}_y=_a\widetilde{SN}^{n\beta}_x$.

\begin{defn}\label{def6} Let $\beta$ be a hesitant fuzzy number,  $F:E\rightarrow HF(U)$, $A\sqsubset E$.

(1) $\overline{SN}^{p\beta}_x=\{y\in U: \beta\subset_p \widetilde{SN}^{p\beta}_x(y)\}$ is called the hesitant soft $p\beta$-neighborhood of $x$ in $T_p=(U,F,A)_{\beta}$.

(2) $\overline{SN}^{a\beta}_x=\{y\in U: \beta\subset_a \widetilde{SN}^{a\beta}_x(y)\}$ is called the hesitant soft $a\beta$-neighborhood of $x$ in $T_a=(U,F,A)_{\beta}$.

(3) $\overline{SN}^{m\beta}_x=\{y\in U: \beta\subset_m \widetilde{SN}^{m\beta}_x(y)\}$ is called the hesitant soft $m\beta$-neighborhood of $x$ in $T_m=(U,F,A)_{\beta}$.

(4) $\overline{SN}^{s\beta}_x=\{y\in U: \beta\subset_s \widetilde{SN}^{s\beta}_x(y)\}$ is called the hesitant soft $s\beta$-neighborhood of $x$ in $T_s=(U,F,A)_{\beta}$.

(5) $\overline{SN}^{t\beta}_x=\{y\in U: \beta\subset_t \widetilde{SN}^{t\beta}_x(y)\}$ is called the hesitant soft $t\beta$-neighborhood of $x$ in $T_t=(U,F,A)_{\beta}$.

(6) $\overline{SN}^{n\beta}_x=\{y\in U: \beta\subset_n \widetilde{SN}^{n\beta}_x(y)\}$ is called the hesitant soft $n\beta$-neighborhood of $x$ in $T_n=(U,F,A)_{\beta}$.
\end{defn}

\begin{exam}\label{exam319} The triple $(U,F,E)$ are shown in Table 2. Let $\beta=\{0.5,0.4,0.3\}$.

We take the results in Example \ref{examp1} to analyse the following cases.

(1)  Let $A=\{e_1,e_2\}$, $T_p=(U,F,A)_{\beta}$, $\widetilde{SN}^{p\beta}_{x_1}=F(e_1)\cap F(e_2)=\frac{0.5,0.4,0.3}{x_1}+\frac{0.4,0.3,0.2}{x_2}+\frac{0.5,0.3,0.3}{x_3}+\frac{1,1,1,1,1,0.2}{x_4}+\frac{0.7,0.3,0.2}{x_5}$.  $\beta\not\subset_p\widetilde{SN}^{p\beta}_{x_1}(x_2)$, then $x_2\not\in\overline{SN}^{p\beta}_{x_1}$. $\beta\subset_p\widetilde{SN}^{p\beta}_{x_1}(x_i)$ for $i=1,3,4,5$, then $\overline{SN}^{p\beta}_{x_1}=\{x_1,x_3,x_4,x_5\}$.

(2)  Let $A=\{e_1,e_2\}$, $T_a=(U,F,A)_{\beta}$, $\widetilde{SN}^{a\beta}_{x_1}=F(e_1)\cap F(e_2)$. $\beta\subset_a\widetilde{SN}^{a\beta}_{x_1}(x_1)$ and $\beta\subset_a\widetilde{SN}^{a\beta}_{x_1}(x_3)$, then  $\overline{SN}^{a\beta}_{x_1}=\{x_1,x_3\}$.

(3)  Let $A=\{e_2,e_3\}$, $T_m=(U,F,A)_{\beta}$, $\widetilde{SN}^{m\beta}_{x_1}=F(e_2)\cap F(e_3)=\frac{0.7,0.5,0.2}{x_1}+\frac{0.4,0.4,0.3,0.2}{x_2}+\frac{0.5,0.5,0.3,0.3}{x_3}+\frac{0.5,0.4}{x_4}+\frac{0.6,0.5,0.2}{x_5}$. $\beta\not\subset_m\widetilde{SN}^{m\beta}_{x_1}(x_2)$, then $x_2\not\in\overline{SN}^{m\beta}_{x_1}$. $\beta\subset_m\widetilde{SN}^{m\beta}_{x_1}(x_i)$ for $i=1,3,4,5$, then $\overline{SN}^{m\beta}_{x_1}=\{x_1,x_3,x_4,x_5\}$.

(4) Let $A=\{e_3,e_4\}$, $T_s=(U,F,A)_{\beta}$, $\widetilde{SN}^{s\beta}_{x_1}=F(e_4)=\frac{0.8,0.7}{x_1}+\frac{0.6,0.1}{x_2}+\frac{0.9,0.8,0.2}{x_3}+\frac{0.6,0.5}{x_4}+\frac{0.6,0.6}{x_5}$. $\beta\subset_s\widetilde{SN}^{s\beta}_{x_1}(x_1)$, $\beta\subset_s\widetilde{SN}^{s\beta}_{x_1}(x_4)$ and $\beta\subset_s\widetilde{SN}^{s\beta}_{x_1}(x_5)$, then $\overline{SN}^{s\beta}_{x_1}=\{x_1,x_4,x_5\}$.

(5) Let $A=\{e_4,e_5\}$, $T_t=(U,F,A)_{\beta}$, $\widetilde{SN}^{t\beta}_{x_2}=F(e_5)=\frac{0.8,0.7,0.7}{x_1}+\frac{0.6,0.5,0.4,0.3}{x_2}+
\frac{0.9,0.8}{x_3}+\frac{0.7,0.6,0.5,0.1}{x_4}+\frac{0.8,0.7,0.6,0.1}{x_5}$. $\beta\subset_t\widetilde{SN}^{t\beta}_{x_2}(x_2)$, $\beta\subset_t\widetilde{SN}^{t\beta}_{x_2}(x_4)$ and $\beta\subset_t\widetilde{SN}^{t\beta}_{x_2}(x_5)$, then $\overline{SN}^{t\beta}_{x_2}=\{x_2,x_4,x_5\}$.

$\widetilde{SN}^{t\beta}_{x_1}=Null$, then $\overline{SN}^{t\beta}_{x_1}=Null$.

(6) Let $A=\{e_5,e_6\}$, $T_n=(U,F,A)_{\beta}$, $\widetilde{SN}^{n\beta}_{x_1}=F(e_5)\cap F(e_6)=\frac{0.6,0.6}{x_1}+\frac{0.6,0.5,0.5,0.4,0.3}{x_2}+
\frac{0.7,0.6}{x_3}+\frac{0.7,0.7,0.6,0.5,0.1}{x_4}+\frac{0.7,0.7,0.6,0.6,0.1}{x_5}$. $\beta\subset_n\widetilde{SN}^{n\beta}_{x_1}(x_1)$ and $\beta\subset_n\widetilde{SN}^{n\beta}_{x_1}(x_3)$, then $\overline{SN}^{n\beta}_{x_1}=\{x_1,x_3\}$.

\end{exam}

\begin{pro}\label{pro320} Let $\beta$ be a hesitant fuzzy number,  the following statements hold for all $x\in U$.

(1) $x\in\overline{SN}^{p\beta}_x$ for each $T_p$.

(2) $x\in\overline{SN}^{a\beta}_x$ for each $T_a$.

(3) $x\in\overline{SN}^{t\beta}_x$ for each $T_t$.

(4) $x\in\overline{SN}^{n\beta}_x$ for each $T_n$.
\end{pro}

{\bf\slshape Proof} (1) By Proposition \ref{pro7} (1), $\beta\subset_p\widetilde{SN}^{p\beta}_x(x)$, then $x\in\overline{SN}^{p\beta}_x$.

(2) By Proposition \ref{pro7} (2), $\beta\subset_a\widetilde{SN}^{a\beta}_x(x)$, then $x\in\overline{SN}^{a\beta}_x$.

(3)  By Proposition \ref{pro7} (3), $\beta\subset_t\widetilde{SN}^{t\beta}_x(x)$, then $x\in\overline{SN}^{t\beta}_x$.

(4)  By Proposition \ref{pro7} (4), $\beta\subset_n\widetilde{SN}^{n\beta}_x(x)$, then $x\in\overline{SN}^{n\beta}_x$.$\blacksquare$

\begin{pro}\label{pro321} Let $\beta$ be a hesitant fuzzy number, the following statements hold for all $x,y\in U$.

(1) For each $T_p$, $y\in\overline{SN}^{p\beta}_x$ if and only if $\overline{SN}^{p\beta}_y\sqsubset\overline{SN}^{p\beta}_x$.

(2) For each $T_p$, $x\in\overline{SN}^{p\beta}_y$ and $y\in\overline{SN}^{p\beta}_x$ if and only if $\overline{SN}^{p\beta}_x=\overline{SN}^{p\beta}_y$.

(3) For each $T_a$, $y\in\overline{SN}^{a\beta}_x$ if and only if $\overline{SN}^{a\beta}_y\sqsubset\overline{SN}^{a\beta}_x$.

(4) For each $T_a$, $x\in\overline{SN}^{a\beta}_y$ and $y\in\overline{SN}^{a\beta}_x$ if and only if $\overline{SN}^{a\beta}_x=\overline{SN}^{a\beta}_y$.
\end{pro}

{\bf\slshape Proof} (1) By Proposition \ref{pro8} (1), $\beta\subset_p \widetilde{SN}^{p\beta}_x(y)$ if and only if $\widetilde{SN}^{p\beta}_y\subset_p \widetilde{SN}^{p\beta}_x$.

If $y\in\overline{SN}^{p\beta}_x$, i.e., $\beta\subset_p \widetilde{SN}^{p\beta}_x(y)$, then $\widetilde{SN}^{p\beta}_y\subset_p \widetilde{SN}^{p\beta}_x$, that is $\widetilde{SN}^{p\beta}_y(z)\subset_p \widetilde{SN}^{p\beta}_x(z)$ for each $z\in U$. If $\beta\subset_p\widetilde{SN}^{p\beta}_y(z)$, then $\beta\subset_p\widetilde{SN}^{p\beta}_x(z)$. It means that if $z\in\overline{SN}^{p\beta}_y$ then $z\in\overline{SN}^{p\beta}_x$, i.e., $\overline{SN}^{p\beta}_y\sqsubset\overline{SN}^{p\beta}_x$.

On the other hand, we have $y\in\overline{SN}^{p\beta}_y$ by the results of Proposition \ref{pro320} (1). If $\overline{SN}^{p\beta}_y\sqsubset\overline{SN}^{p\beta}_x$, then $y\in\overline{SN}^{p\beta}_x$.

(2) By the results of (1), $x\in\overline{SN}^{p\beta}_y$ and $y\in\overline{SN}^{p\beta}_x$ if and only if $\overline{SN}^{p\beta}_x\sqsubset\overline{SN}^{p\beta}_y$ and $\overline{SN}^{p\beta}_y\sqsubset\overline{SN}^{p\beta}_x$, i.e., $\overline{SN}^{p\beta}_y=\overline{SN}^{p\beta}_x$.

(3) By Proposition \ref{pro8} (3) and Proposition \ref{pro320} (2), it can be proved similarly to (1).

(4) By the results of (3), it can be proved similarly to (2).$\blacksquare$

\begin{pro}\label{pro322} Let $\beta$ be a hesitant fuzzy number,  the following statements hold for all $x,y,z\in U$.

(1) For each $T_p$, $y\in\overline{SN}^{p\beta}_x$ and $z\in\overline{SN}^{p\beta}_y$, then $z\in\overline{SN}^{p\beta}_x$.

(2) For each $T_a$, $y\in\overline{SN}^{a\beta}_x$ and $z\in\overline{SN}^{a\beta}_y$, then $z\in\overline{SN}^{a\beta}_x$.

(3) For each $T_n$, $y\in\overline{SN}^{n\beta}_x$ and $z\in\overline{SN}^{n\beta}_y$, then $z\in\overline{SN}^{n\beta}_x$.

\end{pro}

{\bf\slshape Proof} (1) $y\in\overline{SN}^{p\beta}_x$ and $z\in\overline{SN}^{p\beta}_y$, i.e., $\beta\subset_p \widetilde{SN}^{p\beta}_x(y)$ and $\beta\subset_p \widetilde{SN}^{p\beta}_y(z)$. By Proposition \ref{prop1} (1), if $\beta\subset_p\widetilde{SN}^{p\beta}_x(y)$ and $\beta\subset_p\widetilde{SN}^{p\beta}_y(z)$, then $\beta\subset_p\widetilde{SN}^{p\beta}_x(z)$, i.e., $z\in\overline{SN}^{p\beta}_x$.

(2) By Proposition \ref{prop1} (2),  it can be proved similarly to (1).

(3) By Proposition \ref{prop1} (3),  it can be proved similarly to (1).$\blacksquare$

\begin{pro}\label{pro322n1} Let $\beta$ be a hesitant fuzzy number,  the following statements hold for all $x,y\in U$.

(1) For each $T_p$, $\widetilde{SN}^{p\beta}_y\subset_p \widetilde{SN}^{p\beta}_x$ if and only if $\overline{SN}^{p\beta}_y\sqsubset\overline{SN}^{p\beta}_x$.

(2) For each $T_p$, $\overline{SN}^{p\beta}_x=\overline{SN}^{p\beta}_y$ if and only if $\widetilde{SN}^{p\beta}_x=_p \widetilde{SN}^{p\beta}_y$.

(3) For each $T_a$, $\widetilde{SN}^{a\beta}_y\subset_a \widetilde{SN}^{a\beta}_x$ if and only if $\overline{SN}^{a\beta}_y\sqsubset\overline{SN}^{a\beta}_x$.

(4) For each $T_a$, $\overline{SN}^{a\beta}_x=\overline{SN}^{a\beta}_y$ if and only if $\widetilde{SN}^{a\beta}_x=_a \widetilde{SN}^{a\beta}_y$.
\end{pro}

{\bf\slshape Proof} (1)  $y\in\overline{SN}^{p\beta}_x$ i.e., $\beta\subset_p \widetilde{SN}^{p\beta}_x(y)$. By Proposition \ref{pro321} (1), $y\in\overline{SN}^{p\beta}_x$ if and only if $\overline{SN}^{p\beta}_y\sqsubset\overline{SN}^{p\beta}_x$. By Proposition \ref{pro8} (1), $\beta\subset_p \widetilde{SN}^{p\beta}_x(y)$ if and only if $\widetilde{SN}^{p\beta}_y\subset_p \widetilde{SN}^{p\beta}_x$. Then,  $\widetilde{SN}^{p\beta}_y\subset_p \widetilde{SN}^{p\beta}_x$ if and only if $\overline{SN}^{p\beta}_y\sqsubset\overline{SN}^{p\beta}_x$.

(2) By the results of (1), $\overline{SN}^{p\beta}_y\sqsubset\overline{SN}^{p\beta}_x$ and $\overline{SN}^{p\beta}_x\sqsubset\overline{SN}^{p\beta}_y$ if and only if $\widetilde{SN}^{p\beta}_y\subset_p \widetilde{SN}^{p\beta}_x$ and $\widetilde{SN}^{p\beta}_x\subset_p \widetilde{SN}^{p\beta}_y$. It means that $\overline{SN}^{p\beta}_y=\overline{SN}^{p\beta}_x$ if and only if $\widetilde{SN}^{p\beta}_x=_p \widetilde{SN}^{p\beta}_y$.

(3) By Proposition \ref{pro321} (3) and Proposition \ref{pro8} (3),  it can be proved similarly to (1).

(4) By the results of (3),   it can be proved similarly to (2).$\blacksquare$

\begin{pro}\label{pro322n} Let $\beta$ be a hesitant fuzzy number,  the following statements hold for all $x,y\in U$.

(1)   For each $T_p$, $\overline{\widetilde{SN}^{p\beta}_x\cup\widetilde{SN}^{p\beta}_y}^{p\beta}=\overline{SN}^{p\beta}_x\sqcup\overline{SN}^{p\beta}_y$.

(2)  For each $T_p$,  $\overline{\widetilde{SN}^{p\beta}_x\cap\widetilde{SN}^{p\beta}_y}^{p\beta}=\overline{SN}^{p\beta}_x\sqcap\overline{SN}^{p\beta}_y$.

(3)   For each $T_a$, $\overline{SN}^{a\beta}_x\sqcup\overline{SN}^{a\beta}_y\sqsubset\overline{\widetilde{SN}^{a\beta}_x\cup\widetilde{SN}^{a\beta}_y}^{a\beta}$.

(4)  For each $T_a$,  $\overline{\widetilde{SN}^{a\beta}_x\cap\widetilde{SN}^{a\beta}_y}^{a\beta}=\overline{SN}^{a\beta}_x\sqcap\overline{SN}^{a\beta}_y$.

(5)  For each $T_t$, $\overline{SN}^{t\beta}_x\sqcap\overline{SN}^{t\beta}_y\sqsubset\overline{\widetilde{SN}^{t\beta}_x\cap\widetilde{SN}^{t\beta}_y}^{t\beta}$.

(6)  For each $T_n$, $\overline{\widetilde{SN}^{n\beta}_x\cup\widetilde{SN}^{n\beta}_y}^{n\beta}=\overline{SN}^{n\beta}_x\sqcup\overline{SN}^{n\beta}_y$.

(7)  For each $T_n$, $\overline{\widetilde{SN}^{n\beta}_x\cap\widetilde{SN}^{n\beta}_y}^{n\beta}=\overline{SN}^{n\beta}_x\sqcap\overline{SN}^{n\beta}_y$.
\end{pro}

{\bf\slshape Proof} (1)  For each $z\in U$, if $z\in\overline{\widetilde{SN}^{p\beta}_x\cup\widetilde{SN}^{p\beta}_y}^{p\beta}$, then $\beta\subset_p H(z)$, where $\widetilde{SN}^{p\beta}_x\cup\widetilde{SN}^{p\beta}_y=H\in HF(U)$ and $\overline{H}^{p\beta}=\{z:\beta\subset_p H(z),z\in U\}$, then $\beta^+\leqslant H^+(z)=\max\{(\widetilde{SN}^{p\beta}_x)^+(z),(\widetilde{SN}^{p\beta}_y)^+(z)\}$. It means that at least one of $\beta\subset_p \widetilde{SN}^{p\beta}_x(z)$ and $\beta\subset_p \widetilde{SN}^{p\beta}_y(z)$ holds, i.e., $z\in\overline{SN}^{p\beta}_x\sqcup\overline{SN}^{p\beta}_y$.

On the other hand, if $z\in\overline{SN}^{p\beta}_x\sqcup\overline{SN}^{p\beta}_y$, i.e., $\beta\subset_p\widetilde{SN}^{p\beta}_x(z)$ or $\beta\subset_p\widetilde{SN}^{p\beta}_y(z)$. By Theorem \ref{thm2.26} (2), $\beta\subset_p(\widetilde{SN}^{p\beta}_x\cup\widetilde{SN}^{p\beta}_y)(z)$, i.e., $z\in\overline{\widetilde{SN}^{p\beta}_x\cup\widetilde{SN}^{p\beta}_y}^{p\beta}$.

(2) By Theorem \ref{thm2.26} (1), $\beta\subset_p\widetilde{SN}^{p\beta}_x(z)$ and $\beta\subset_p\widetilde{SN}^{p\beta}_y(z)$ if and only if  $\beta\subset_p(\widetilde{SN}^{p\beta}_x\cap\widetilde{SN}^{p\beta}_y)(z)$. Then $\overline{\widetilde{SN}^{p\beta}_x\cap\widetilde{SN}^{p\beta}_y}^{p\beta}=\overline{SN}^{p\beta}_x\sqcap\overline{SN}^{p\beta}_y$.

(3)  For each $z\in U$, if $z\in\overline{SN}^{a\beta}_x\sqcup\overline{SN}^{a\beta}_y$, i.e., $\beta\subset_a\widetilde{SN}^{a\beta}_x(z)$ or $\beta\subset_a\widetilde{SN}^{a\beta}_y(z)$. By Theorem \ref{thm2.26} (4), $\beta\subset_a(\widetilde{SN}^{a\beta}_x\cup\widetilde{SN}^{a\beta}_y)(z)$, i.e., $z\in\overline{\widetilde{SN}^{a\beta}_x\cup\widetilde{SN}^{a\beta}_y}^{a\beta}$. Then, $\overline{SN}^{a\beta}_x\sqcup\overline{SN}^{a\beta}_y\sqsubset\overline{\widetilde{SN}^{a\beta}_x\cup\widetilde{SN}^{a\beta}_y}^{a\beta}$.

(4) By Theorem \ref{thm2.26} (3),  it can be proved similarly to (2).

(5) By Theorem \ref{thm2.26} (5), if $\beta\subset_t\widetilde{SN}^{t\beta}_x(z)$ and $\beta\subset_t\widetilde{SN}^{t\beta}_y(z)$, then  $\beta\subset_t(\widetilde{SN}^{t\beta}_x\cap\widetilde{SN}^{t\beta}_y)(z)$. Then $\overline{SN}^{t\beta}_x\sqcap\overline{SN}^{t\beta}_y\sqsubset\overline{\widetilde{SN}^{t\beta}_x\cap\widetilde{SN}^{t\beta}_y}^{t\beta}$.

(6) For all $z\in U$, if $z\in\overline{\widetilde{SN}^{n\beta}_x\cup\widetilde{SN}^{n\beta}_y}^{n\beta}$, then $\beta\subset_n H(z)$,  where $\widetilde{SN}^{n\beta}_x\cup\widetilde{SN}^{n\beta}_y=H\in HF(U)$, $\overline{H}^{n\beta}=\{z:\beta\subset_n H(z),z\in U\}$, then $\beta^+\leqslant H^-(z)=\max\{(\widetilde{SN}^{n\beta}_x)^-(z),(\widetilde{SN}^{n\beta}_y)^-(z)\}$. It means that at least one of $\beta\subset_n \widetilde{SN}^{n\beta}_x(z)$ and $\beta\subset_n \widetilde{SN}^{n\beta}_y(z)$ holds, i.e., $z\in\overline{SN}^{n\beta}_x\sqcup\overline{SN}^{n\beta}_y$.

On the other hand, if $z\in\overline{SN}^{n\beta}_x\sqcup\overline{SN}^{n\beta}_y$, i.e., $\beta\subset_n\widetilde{SN}^{n\beta}_x(z)$ or $\beta\subset_n\widetilde{SN}^{n\beta}_y(z)$. By Theorem \ref{thm2.26} (8), $\beta\subset_n(\widetilde{SN}^{n\beta}_x\cup\widetilde{SN}^{n\beta}_y)(z)$, i.e., $z\in\overline{\widetilde{SN}^{n\beta}_x\cup\widetilde{SN}^{n\beta}_y}^{n\beta}$.

(7) By Theorem \ref{thm2.26} (7),  it can be proved similarly to (2).$\blacksquare$

\begin{pro}\label{pro323} Let $\beta$ be a hesitant fuzzy number,  $N_0<\infty$, $\Lambda$ is a random set of indicators,  the following statements hold for all $x_i,x_v\in U$.

(1)  For each $T_p$, $\sqcup_{i=1}^{N_0}\overline{SN}_{x_i}^{p\beta}=\overline{\bigcup_{i=1}^{N_0}\widetilde{SN}_{x_i}^{p\beta}}^{p\beta}$.

(2)  For each $T_p$, $\sqcap_{i=1}^{N_0}\overline{SN}_{x_i}^{p\beta}=\overline{\bigcap_{i=1}^{N_0}\widetilde{SN}_{x_i}^{p\beta}}^{p\beta}$.

(3)  For each $T_p$, $\sqcup_{v\in\Lambda}\overline{SN}_{x_v}^{p\beta}\sqsubset\overline{\bigcup_{v\in\Lambda}\widetilde{SN}_{x_v}^{p\beta}}^{p\beta}$.

(4)  For each $T_p$, $\sqcap_{v\in\Lambda}\overline{SN}_{x_v}^{p\beta}=\overline{\bigcap_{v\in\Lambda}\widetilde{SN}_{x_v}^{p\beta}}^{p\beta}$.

(5)  For each $T_a$, $\sqcup_{i=1}^{N_0}\overline{SN}_{x_i}^{a\beta}\sqsubset\overline{\bigcup_{i=1}^{N_0}\widetilde{SN}_{x_i}^{a\beta}}^{a\beta}$.

(6)  For each $T_a$, $\sqcap_{i=1}^{N_0}\overline{SN}_{x_i}^{a\beta}=\overline{\bigcap_{i=1}^{N_0}\widetilde{SN}_{x_i}^{a\beta}}^{a\beta}$.

(7)  For each $T_a$, $\sqcup_{v\in\Lambda}\overline{SN}_{x_v}^{a\beta}\sqsubset\overline{\bigcup_{v\in\Lambda}\widetilde{SN}_{x_v}^{a\beta}}^{a\beta}$.

(8)  For each $T_a$, $\sqcap_{v\in\Lambda}\overline{SN}_{x_v}^{a\beta}=\overline{\bigcap_{v\in\Lambda}\widetilde{SN}_{x_v}^{a\beta}}^{a\beta}$.

(9) For each $T_t$, $\sqcap_{i=1}^{N_0}\overline{SN}_{x_i}^{t\beta}\sqsubset\overline{\bigcap_{i=1}^{N_0}\widetilde{SN}_{x_i}^{t\beta}}^{t\beta}$.

(10) For each $T_t$, $\sqcap_{v\in\Lambda}\overline{SN}_{x_v}^{t\beta}\sqsubset\overline{\bigcap_{v\in\Lambda}\widetilde{SN}_{x_v}^{t\beta}}^{t\beta}$.

(11) For each $T_n$, $\sqcup_{i=1}^{N_0}\overline{SN}_{x_i}^{n\beta}=\overline{\bigcup_{i=1}^{N_0}\widetilde{SN}_{x_i}^{n\beta}}^{n\beta}$.

(12) For each $T_n$, $\sqcap_{i=1}^{N_0}\overline{SN}_{x_i}^{n\beta}=\overline{\bigcap_{i=1}^{N_0}\widetilde{SN}_{x_i}^{n\beta}}^{n\beta}$.

(13) For each $T_n$, $\sqcup_{v\in\Lambda}\overline{SN}_{x_v}^{n\beta}\sqsubset\overline{\bigcup_{v\in\Lambda}\widetilde{SN}_{x_v}^{n\beta}}^{n\beta}$.

(14) For each $T_n$, $\sqcap_{v\in\Lambda}\overline{SN}_{x_v}^{n\beta}=\overline{\bigcap_{v\in\Lambda}\widetilde{SN}_{x_v}^{n\beta}}^{n\beta}$.
\end{pro}

{\bf\slshape Proof} (1) By the result of Proposition \ref{pro322n} (1), it can be proved by using mathematical induction.

(2) By the result of Proposition \ref{pro322n} (2), it can be proved by using mathematical induction.

(3) For each $z\in U$, if $z\in\sqcup_{v\in\Lambda}\overline{SN}_{x_v}^{p\beta}$, then there exists at least one $v'\in\Lambda$ such that $z\in\overline{SN}_{x_{v'}}^{p\beta}$, i.e., $\beta\subset_p(\widetilde{SN}_{x_{v'}}^{p\beta})(z)$. By Theorem \ref{thm2.26} (2),
$\beta\subset_p(\bigcup_{v\in\Lambda}\widetilde{SN}_{x_v}^{p\beta})(z)$, i.e., $z\in\overline{\bigcup_{v\in\Lambda}\widetilde{SN}_{x_v}^{p\beta}}^{p\beta}$. Then,  $\sqcup_{v\in\Lambda}\overline{SN}_{x_v}^{p\beta}\sqsubset\overline{\bigcup_{v\in\Lambda}\widetilde{SN}_{x_v}^{p\beta}}^{p\beta}$.

(4) For each $z\in U$,  $z\in\sqcap_{v\in\Lambda}\overline{SN}_{x_v}^{p\beta}$ if and only if $z\in\overline{SN}_{x_v}^{p\beta}$ for all $v\in\Lambda$, i.e., $\beta\subset_p(\widetilde{SN}_{x_{v}}^{p\beta})(z)$ holds for all $v\in\Lambda$.

By Theorem \ref{thm2.26} (1),  $\beta\subset_p(\bigcap_{v\in\Lambda}\widetilde{SN}_{x_v}^{p\beta})(z)$ if and only if $\beta\subset_p(\widetilde{SN}_{x_{v}}^{p\beta})(z)$ for all $v\in\Lambda$, i.e., $\sqcap_{v\in\Lambda}\overline{SN}_{x_v}^{p\beta}=\overline{\bigcap_{v\in\Lambda}\widetilde{SN}_{x_v}^{p\beta}}^{p\beta}$.

(5) By the result of Proposition \ref{pro322n} (3), it can be proved by using mathematical induction.

(6) By the result of Proposition \ref{pro322n} (4), it can be proved by using mathematical induction.

(7)  For each $z\in U$, if $z\in\sqcup_{v\in\Lambda}\overline{SN}_{x_v}^{a\beta}$, then there exists at least one $v'\in\Lambda$ such that $z\in\overline{SN}_{x_{v'}}^{a\beta}$, i.e., $\beta\subset_a(\widetilde{SN}_{x_{v'}}^{a\beta})(z)$. By Theorem \ref{thm2.26} (4),
$\beta\subset_a(\bigcup_{v\in\Lambda}\widetilde{SN}_{x_v}^{a\beta})(z)$, i.e., $z\in\overline{\bigcup_{v\in\Lambda}\widetilde{SN}_{x_v}^{a\beta}}^{a\beta}$. Then,  $\sqcup_{v\in\Lambda}\overline{SN}_{x_v}^{a\beta}\sqsubset\overline{\bigcup_{v\in\Lambda}\widetilde{SN}_{x_v}^{a\beta}}^{a\beta}$.

(8) For each $z\in U$,  $z\in\sqcap_{v\in\Lambda}\overline{SN}_{x_v}^{a\beta}$ if and only if $z\in\overline{SN}_{x_v}^{a\beta}$ for all $v\in\Lambda$, i.e., $\beta\subset_a(\widetilde{SN}_{x_{v}}^{a\beta})(z)$ holds for all $v\in\Lambda$.

By Theorem \ref{thm2.26} (3),  $\beta\subset_a(\bigcap_{v\in\Lambda}\widetilde{SN}_{x_v}^{a\beta})(z)$ if and only if $\beta\subset_a(\widetilde{SN}_{x_{v}}^{a\beta})(z)$ for all $v\in\Lambda$, i.e., $\sqcap_{v\in\Lambda}\overline{SN}_{x_v}^{a\beta}=\overline{\bigcap_{v\in\Lambda}\widetilde{SN}_{x_v}^{a\beta}}^{a\beta}$.

(9) By the conclusion of Proposition \ref{pro322n} (5), it can be proved by using mathematical induction.

(10) For all $z\in U$, if $z\in\sqcap_{v\in\Lambda}\overline{SN}_{x_v}^{t\beta}$, then $z\in\overline{SN}_{x_v}^{t\beta}$ for all $v\in\Lambda$, then $\beta\subset_t\widetilde{SN}_{x_v}^{t\beta}(z)$ holds for all $v\in\Lambda$. By Theorem \ref{thm2.26} (5), $\beta\subset_t(\bigcap_{v\in\Lambda}\widetilde{SN}_{x_v}^{t\beta})(z)$, i.e.,  $z\in\overline{\bigcap_{v\in\Lambda}\widetilde{SN}_{x_v}^{t\beta}}^{t\beta}$. Then, $\sqcap_{v\in\Lambda}\overline{SN}_{x_v}^{t\beta}\sqsubset\overline{\bigcap_{v\in\Lambda}\widetilde{SN}_{x_v}^{t\beta}}^{t\beta}$.

(11) By the conclusion of Proposition \ref{pro322n} (6), it can be proved by using mathematical induction.

(12) By the conclusion of Proposition \ref{pro322n} (7), it can be proved by using mathematical induction.

(13)  For each $z\in U$, if $z\in\sqcup_{v\in\Lambda}\overline{SN}_{x_v}^{n\beta}$, then there exists at least one $v'\in\Lambda$ such that $z\in\overline{SN}_{x_{v'}}^{n\beta}$, i.e., $\beta\subset_n(\widetilde{SN}_{x_{v'}}^{n\beta})(z)$. By Theorem \ref{thm2.26} (8),
$\beta\subset_n(\bigcup_{v\in\Lambda}\widetilde{SN}_{x_v}^{n\beta})(z)$, i.e., $z\in\overline{\bigcup_{v\in\Lambda}\widetilde{SN}_{x_v}^{n\beta}}^{n\beta}$. Then $\sqcup_{v\in\Lambda}\overline{SN}_{x_v}^{n\beta}\sqsubset\overline{\bigcup_{v\in\Lambda}\widetilde{SN}_{x_v}^{n\beta}}^{n\beta}$.

(14) For all $z\in U$,  $z\in\sqcap_{v\in\Lambda}\overline{SN}_{x_v}^{n\beta}$ if and only if $z\in\overline{SN}_{x_v}^{n\beta}$ for all $v\in\Lambda$, i.e., $\beta\subset_n(\widetilde{SN}_{x_{v}}^{n\beta})(z)$ holds for all $v\in\Lambda$.

By Theorem \ref{thm2.26} (7),  $\beta\subset_n(\bigcap_{v\in\Lambda}\widetilde{SN}_{x_v}^{n\beta})(z)$ if and only if $\beta\subset_n(\widetilde{SN}_{x_{v}}^{n\beta})(z)$ for all $v\in\Lambda$, i.e., $\sqcap_{v\in\Lambda}\overline{SN}_{x_v}^{n\beta}=\overline{\bigcap_{v\in\Lambda}\widetilde{SN}_{x_v}^{n\beta}}^{n\beta}$.$\blacksquare$\\

Let $\beta=\{0.5,0.4,0.3\}$.

If $(\widetilde{SN}_{x_i}^{p\beta})^+(z)=0.5(1-\frac{1}{i})$, where $i$ is a positive integer. $(\bigcup_{i=1}^{\infty}\widetilde{SN}_{x_i}^{p\beta})^+(z)=\lim\limits_{i\to\infty}0.5(1-\frac{1}{i})=0.5\geqslant\beta^+(z)$. Then $z\in\overline{\bigcup_{i=1}^{\infty}\widetilde{SN}_{x_i}^{p\beta}}^{p\beta}$. However, $z\not\in\overline{SN}_{x_i}^{p\beta}$ for every positive integer $i$. Hence, the (3) of Proposition \ref{pro323}   cannot take the equal sign.

If $(\widetilde{SN}_{x_i}^{n\beta})^-(z)=0.5(1-\frac{1}{i})$, where $i$ is a positive integer. $(\bigcup_{i=1}^{\infty}\widetilde{SN}_{x_i}^{n\beta})^-(z)=\lim\limits_{i\to\infty}0.5(1-\frac{1}{i})=0.5\geqslant\beta^+(z)$. Then $z\in\overline{\bigcup_{i=1}^{\infty}\widetilde{SN}_{x_i}^{n\beta}}^{n\beta}$. However, $z\not\in\overline{SN}_{x_i}^{n\beta}$ for every positive integer $i$. Hence, the (13) of Proposition \ref{pro323}   cannot take the equal sign.

If $\widetilde{SN}_{x_1}^{a\beta}(z)=\{0.5,0.1\}$ and $\widetilde{SN}_{x_2}^{a\beta}(z)=\{0.4,0.3\}$, it is obvious $z\not\in\overline{SN}_{x_1}^{a\beta}$ and $z\not\in\overline{SN}_{x_2}^{a\beta}$. $\beta\subset_a(\widetilde{SN}_{x_1}^{a\beta}\cup\widetilde{SN}_{x_2}^{a\beta})(z)=\{0.5,0.4,0.3\}$, then $z\in\overline{\bigcup_{i=1}^2\widetilde{SN}_{x_i}^{a\beta}}^{a\beta}$. Hence, the (5) and (7) of Proposition \ref{pro323}  cannot take the equal sign.

If $\widetilde{SN}_{x_1}^{t\beta}(z)=\{0.5,0.3,0.1\}$ and $\widetilde{SN}_{x_2}^{t\beta}(z)=\{0.5,0.4,0.1\}$, it is obvious $z\not\in\overline{SN}_{x_1}^{t\beta}$ and $z\not\in\overline{SN}_{x_2}^{t\beta}$. $\beta\subset_t(\widetilde{SN}_{x_1}^{t\beta}\cap\widetilde{SN}_{x_2}^{t\beta})(z)=\{0.5,0.5,0.4,0.3,0.1,0.1\}$, $z\in\overline{\bigcap_{i=1}^2\widetilde{SN}_{x_i}^{t\beta}}^{t\beta}$. Hence, the (10) of Proposition \ref{pro323} cannot take the equal sign.

\section{Hesitant fuzzy soft $\beta$-covering based rough sets}

Let's first review the concept of rough sets. Let $R$ be an equivalence relation on $U$. Then a pair $(U, R)$ is called
a Pawlak approximation space \cite{PALA82}. Based on $(U, R)$, two rough approximations can be defined as  $R_-(X)=\{x\in U:[x]_R\sqsubset X\}$ and $R_+(X)=\{x\in U:[x]_R\sqcap X\neq\emptyset\}$, where $X\sqsubset U$. $R_-(X)$ and $R_+(X)$ are called the Pawlak lower approximation and the
Pawlak upper approximation of $X$, respectively. The Pawlak boundary region of $X$ is defined as $R_B(X)=R_+(X)-R_-(X)$. It is easy to know that $R_-(X)\sqsubset X\sqsubset R_+(X)$.

\begin{defn}  Let $\beta$ be a hesitant fuzzy number, $X\in HF(U)$, $x,y\in U$.

(1) $\widetilde{SN}_-^{p\beta}(X)(x)=\bigcap\limits_{y\in U}\{(\widetilde{SN}_x^{p\beta})^c(y)\cup X(y)\}$ and $\widetilde{SN}_+^{p\beta}(X)(x)=\bigcup\limits_{y\in U}\{\widetilde{SN}_x^{p\beta}(y)\cap X(y)\}$ are the $p$-lower approximation and $p$-upper approximation of the hesitant fuzzy set $X$ in $T_p=(U,F,A)_{\beta}$, respectively.

(2) $\widetilde{SN}_-^{a\beta}(X)(x)=\bigcap\limits_{y\in U}\{(\widetilde{SN}_x^{a\beta})^c(y)\cup X(y)\}$ and $\widetilde{SN}_+^{a\beta}(X)(x)=\bigcup\limits_{y\in U}\{\widetilde{SN}_x^{a\beta}(y)\cap X(y)\}$ are the $a$-lower approximation and $a$-upper approximation of the hesitant fuzzy set $X$ in $T_a=(U,F,A)_{\beta}$, respectively.

(3) $\widetilde{SN}_-^{m\beta}(X)(x)=\bigcap\limits_{y\in U}\{(\widetilde{SN}_x^{m\beta})^c(y)\cup X(y)\}$ and $\widetilde{SN}_+^{m\beta}(X)(x)=\bigcup\limits_{y\in U}\{\widetilde{SN}_x^{m\beta}(y)\cap X(y)\}$ are the $m$-lower approximation and $m$-upper approximation of the hesitant fuzzy set $X$ in $T_m=(U,F,A)_{\beta}$, respectively.

(4) $\widetilde{SN}_-^{s\beta}(X)(x)=\bigcap\limits_{y\in U}\{(\widetilde{SN}_x^{s\beta})^c(y)\cup X(y)\}$ and $\widetilde{SN}_+^{s\beta}(X)(x)=\bigcup\limits_{y\in U}\{\widetilde{SN}_x^{s\beta}(y)\cap X(y)\}$ are the $s$-lower approximation and $s$-upper approximation of the hesitant fuzzy set $X$ in $T_s=(U,F,A)_{\beta}$, respectively.

(5) $\widetilde{SN}_-^{t\beta}(X)(x)=\bigcap\limits_{y\in U}\{(\widetilde{SN}_x^{t\beta})^c(y)\cup X(y)\}$ and $\widetilde{SN}_+^{t\beta}(X)(x)=\bigcup\limits_{y\in U}\{\widetilde{SN}_x^{t\beta}(y)\cap X(y)\}$ are the $t$-lower approximation and $t$-upper approximation of the hesitant fuzzy set $X$ in $T_t=(U,F,A)_{\beta}$, respectively.

(6) $\widetilde{SN}_-^{n\beta}(X)(x)=\bigcap\limits_{y\in U}\{(\widetilde{SN}_x^{n\beta})^c(y)\cup X(y)\}$ and $\widetilde{SN}_+^{n\beta}(X)(x)=\bigcup\limits_{y\in U}\{\widetilde{SN}_x^{n\beta}(y)\cap X(y)\}$ are the $n$-lower approximation and $n$-upper approximation of the hesitant fuzzy set $X$ in $T_n=(U,F,A)_{\beta}$, respectively.
\end{defn}

\begin{exam}\label{examp2} The triple $(U,F,E)$ are shown in Table 2. Let $\beta=\{0.5,0.3,0.2\}$, $X=\frac{0.2,0.1}{x_1}+\frac{0.2,0.1}{x_2}+\frac{0.1,0.1}{x_3}+\frac{0.2,0.1}{x_4}+\frac{0.2,0.2}{x_5}$.

(1) Let $A=\{e_6,e_7\}$, $\bigcup\limits_{e\in A}F(e)=\frac{0.6,0.6}{x_1}+\frac{0.7,0.5}{x_2}+\frac{0.7,0.6}{x_3}+\frac{0.8,0.7}{x_4}+\frac{0.7,0.6}{x_5}$,  $\beta\subset_p(\bigcup\limits_{e\in A}F(e))(x)$ for all $x\in U$. $T_p=(U,F,A)_{\beta}$ is a hesitant fuzzy soft $p\beta$-covering approximation space.

In this $T_p$, $\beta\subset_p F(e_6)(x_1)$ and $\beta\not\subset_p F(e_7)(x_1)$, then $\widetilde{SN}^{p\beta}_{x_1}=F(e_6)=\frac{0.6,0.6}{x_1}+\frac{0.7,0.5}{x_2}+\frac{0.7,0.6}{x_3}+\frac{0.8,0.7}{x_4}+\frac{0.7,0.6}{x_5}$. $(\widetilde{SN}^{p\beta}_{x_1})^c=\frac{0.4,0.4}{x_1}+\frac{0.5,0.3}{x_2}+\frac{0.4,0.3}{x_3}+\frac{0.3,0.2}{x_4}+\frac{0.4,0.3}{x_5}$.

$\widetilde{SN}_-^{p\beta}(X)(x_1)=\bigcap\limits_{x_i\in U}\{(\widetilde{SN}_{x_1}^{p\beta})^c(x_i)\cup X(x_i)\}=\{(\widetilde{SN}_{x_1}^{p\beta})^c(x_1)\cup X(x_1)\}\cap\{(\widetilde{SN}_{x_1}^{p\beta})^c(x_2)\cup X(x_2)\}\cap\{(\widetilde{SN}_{x_1}^{p\beta})^c(x_3)\cup X(x_3)\}\cap\{(\widetilde{SN}_{x_1}^{p\beta})^c(x_4)\cup X(x_4)\}\cap\{(\widetilde{SN}_{x_1}^{p\beta})^c(x_5)\cup X(x_5)\}=\{0.3,0.3,0.3,0.3,0.2,0.2\}$.

In the same manner above, we can obtain $\widetilde{SN}_-^{p\beta}(X)(x)$ for $x\in U-\{x_1\}$. Then, $\widetilde{SN}_-^{p\beta}(X)=\frac{0.3,0.3,0.3,0.3,0.2,0.2}{x_1}+\frac{0.3,0.3,0.3,0.3,0.2,0.2}{x_2}+\frac{0.3,0.3,0.3,0.3,0.2,0.2}{x_3}+\frac{0.3,0.3,0.3,0.3,0.2,0.2}{x_4}+\frac{0.3,0.3,0.3,0.3,0.2,0.2}{x_5}$.

$\widetilde{SN}_+^{p\beta}(X)(x_1)=\bigcup\limits_{x_i\in U}\{\widetilde{SN}_{x_1}^{p\beta}(x_i)\cap X(x_i)\}=\{\widetilde{SN}_{x_1}^{p\beta}(x_1)\cap X(x_1)\}\cup\{\widetilde{SN}_{x_1}^{p\beta}(x_2)\cap X(x_2)\}\cup\{\widetilde{SN}_{x_1}^{p\beta}(x_3)\cap X(x_3)\}\cup\{\widetilde{SN}_{x_1}^{p\beta}(x_4)\cap X(x_4)\}\cup\{\widetilde{SN}_{x_1}^{p\beta}(x_5)\cap X(x_5)\}=\{0.2,0.2,0.2,0.2,0.2\}$.

In the same manner above, we can obtain $\widetilde{SN}_+^{p\beta}(X)(x)$ for $x\in U-\{x_1\}$. Then, $\widetilde{SN}_+^{p\beta}(X)=\frac{0.2,0.2,0.2,0.2,0.2}{x_1}+\frac{0.2,0.2,0.2,0.2,0.2}{x_2}+\frac{0.2,0.2,0.2,0.2,0.2}{x_3}+\frac{0.2,0.2,0.2,0.2,0.2}{x_4}+\frac{0.2,0.2,0.2,0.2,0.2}{x_5}$.

(2) Let $A=\{e_6,e_7\}$, $\beta\subset_n\bigcup\limits_{e\in A}F(e)$. $T_n=(U,F,A)_{\beta}$ is a hesitant fuzzy soft $n\beta$-covering approximation space.

Furthermore, $\beta\subset_n\bigcup\limits_{e\in A}F(e)$ implies $\beta\subset_s\bigcup\limits_{e\in A}F(e)$, $\beta\subset_m\bigcup\limits_{e\in A}F(e)$ and $\beta\subset_a\bigcup\limits_{e\in A}F(e)$.

$\beta\subset_n F(e_6)(x_1)$ and $\beta\not\subset_n F(e_7)(x_1)$, then $\widetilde{SN}^{n\beta}_{x_1}=F(e_6)$ and $(\widetilde{SN}^{n\beta}_{x_1})^c=(F(e_6))^c$. Then, $\widetilde{SN}_-^{n\beta}(X)(x_1)=\{0.3,0.3,0.3,0.3,0.2,0.2\}$ and  $\widetilde{SN}_+^{n\beta}(X)(x_1)=\{0.2,0.2,0.2,0.2,0.2\}$.

In the same manner, we can calculate $\widetilde{SN}_-^{s\beta}(X)(x_1)=\widetilde{SN}_-^{m\beta}(X)(x_1)=\widetilde{SN}_-^{a\beta}(X)(x_1)=\{0.3,0.3,0.3,0.3,0.2,0.2\}$ and
$\widetilde{SN}_+^{s\beta}(X)(x_1)=\widetilde{SN}_+^{m\beta}(X)(x_1)=\widetilde{SN}_+^{a\beta}(X)(x_1)=\{0.2,0.2,0.2,0.2,0.2\}$.

(3) Let $A=\{e_6,e_8\}$, $\beta\subset_t\bigcup\limits_{e\in A}F(e)$. $T_t=(U,F,A)_{\beta}$ is a hesitant fuzzy soft $t\beta$-covering approximation space.

$\beta\not\subset_t F(e_6)(x_1)$ and $\beta\subset_t F(e_8)(x_1)$, then $\widetilde{SN}^{t\beta}_{x_1}=F(e_8)$ and $(\widetilde{SN}^{t\beta}_{x_1})^c=(F(e_8))^c$.

$\widetilde{SN}_-^{t\beta}(X)(x_1)=\{0.5,0.5,0.5,0.5,0.5,0.5,0.5,0.4,\newline 0.4, 0.4,0.3,0.3,0.3,0.3,0.3,0.2\}$ and $\widetilde{SN}_+^{t\beta}(X)(x_1)=\{0.2,0.2,0.2,0.2,0.2\}$.

\end{exam}

\begin{pro}\label{pro43} Let $\beta$ be a hesitant fuzzy number, $X,Y\in HF(U)$, $x,y\in U$, the following statements hold in $T_p$.

(1)  $\widetilde{SN}_-^{p\beta}(H^U)=_nH^U$, $\widetilde{SN}_+^{p\beta}(H^{\emptyset})=_n H^{\emptyset}$.

(2) $\widetilde{SN}_-^{p\beta}(X^c)=(\widetilde{SN}_+^{p\beta}(X))^c$, $\widetilde{SN}_+^{p\beta}(X^c)=(\widetilde{SN}_-^{p\beta}(X))^c$.

(3)  $\widetilde{SN}_-^{p\beta}(X\cap Y)=_p\widetilde{SN}_-^{p\beta}(X)\cap \widetilde{SN}_-^{p\beta}(Y)$, $\widetilde{SN}_+^{p\beta}(X\cup Y)=_p\widetilde{SN}_+^{p\beta}(X)\cup \widetilde{SN}_+^{p\beta}(Y)$.

(4) If $X\subset_p Y$, then $\widetilde{SN}_-^{p\beta}(X)\subset_p\widetilde{SN}_-^{p\beta}(Y)$, $\widetilde{SN}_+^{p\beta}(X)\subset_p\widetilde{SN}_+^{p\beta}(Y)$.

(5) $\widetilde{SN}_-^{p\beta}(X)\cup\widetilde{SN}_-^{p\beta}(Y)\subset_p\widetilde{SN}_-^{p\beta}(X\cup Y)$, $\widetilde{SN}_+^{p\beta}(X\cap Y)\subset_p\widetilde{SN}_+^{p\beta}(X)\cap\widetilde{SN}_+^{p\beta}(Y)$.

(6) If $(\widetilde{SN}_x^{p\beta})^c(x)\subset_p X(x)\subset_p\widetilde{SN}_x^{p\beta}(x)$ for all $x\in U$, then $\widetilde{SN}_-^{p\beta}(X)\subset_p X(x)\subset_p\widetilde{SN}_+^{p\beta}(X)$.

(7) If $(\widetilde{SN}_x^{p\beta})^c(x)\subset_p X(x)\subset_p\widetilde{SN}_x^{p\beta}(x)$ for all $x\in U$, then $\widetilde{SN}_-^{p\beta}(\widetilde{SN}_-^{p\beta}(X))\subset_p\widetilde{SN}_-^{p\beta}(X)\subset_p X\subset_p\widetilde{SN}_+^{p\beta}(X)\subset_p\widetilde{SN}_+^{p\beta}(\widetilde{SN}_+^{p\beta}(X))$.

(8) If $X\subset_p Y$, $\widetilde{SN}_-^{p\beta}(X)\cup\widetilde{SN}_-^{p\beta}(Y)=_p\widetilde{SN}_-^{p\beta}(X\cup Y)$, $\widetilde{SN}_+^{p\beta}(X\cap Y)=_p\widetilde{SN}_+^{p\beta}(X)\cap\widetilde{SN}_+^{p\beta}(Y)$.
\end{pro}

{\bf\slshape Proof} (1) For each $y\in U$, $(\widetilde{SN}_x^{p\beta})^c(y)\cup H^U(y)=((\widetilde{SN}_x^{p\beta})^c\cup H^U)(y)$. Since $(H^U)^-(y)=1$, then $((\widetilde{SN}_x^{p\beta})^c\cup H^U)^-(y)=1\geqslant (H^U)^+(y)=1$. Then $H^U\subset_n\widetilde{SN}_-^{p\beta}(H^U)$.

Since $A\subset_n H^U$ for all $A\in HF(U)$, then $\widetilde{SN}_-^{p\beta}(H^U)\subset_nH^U$, then $\widetilde{SN}_-^{p\beta}(H^U)=_nH^U$.

$\widetilde{SN}_x^{p\beta}(y)\cap H^{\emptyset}(y)=(\widetilde{SN}_x^{p\beta}\cap H^{\emptyset})(y)$.  Since $(H^{\emptyset})^+(y)=0$, then $(\widetilde{SN}_x^{p\beta}\cap H^{\emptyset})^+(y)=0\leqslant (H^{\emptyset})^-(y)=0$. Then $\widetilde{SN}_+^{p\beta}(H^{\emptyset})\subset_n H^{\emptyset}$.

Since $H^{\emptyset}\subset_n A$ for all $A\in HF(U)$, then $H^{\emptyset}\subset_n\widetilde{SN}_+^{p\beta}(H^{\emptyset})$, then $\widetilde{SN}_+^{p\beta}(H^{\emptyset})=_n H^{\emptyset}$.

(2) We prove the case of $|U|=2$ at first. Suppose $U=\{y_1,y_2\}$.

By the results of Theorem \ref{thm8} (3), (4) and (5), $\widetilde{SN}_-^{p\beta}(X^c)(y_1)=\bigcap\limits_{y_i\in U}\{(\widetilde{SN}_{y_1}^{p\beta})^c(y_i)\cup X^c(y_i)\}=[(\widetilde{SN}_{y_1}^{p\beta})^c(y_1)\cup X^c(y_1)]\cap[(\widetilde{SN}_{y_1}^{p\beta})^c(y_2)\cup X^c(y_2)]=[\widetilde{SN}_{y_1}^{p\beta}(y_1)\cap X(y_1)]^c\cap[\widetilde{SN}_{y_1}^{p\beta}(y_2)\cap X(y_2)]^c=\{[\widetilde{SN}_{y_1}^{p\beta}(y_1)\cap X(y_1)]\cup[\widetilde{SN}_{y_1}^{p\beta}(y_2)\cap X(y_2)]\}^c=(\widetilde{SN}_+^{p\beta}(X))^c(y_1)$.

$\widetilde{SN}_-^{p\beta}(X^c)(y_2)=(\widetilde{SN}_+^{p\beta}(X))^c(y_2)$ can be proved as above.

To sum up,  $\widetilde{SN}_-^{p\beta}(X^c)=(\widetilde{SN}_+^{p\beta}(X))^c$ hold for the case of $|U|=2$. Other cases of $|U|<\infty$ can be proved by using the mathematical induction.

$\widetilde{SN}_+^{p\beta}(X^c)(y_1)=\bigcup\limits_{y_i\in U}\{\widetilde{SN}_{y_1}^{p\beta}(y_i)\cap X^c(y_i)\}=[\widetilde{SN}_{y_1}^{p\beta}(y_1)\cap X^c(y_1)]\cup[\widetilde{SN}_{y_1}^{p\beta}(y_2)\cap X^c(y_2)]=\{[(\widetilde{SN}_{y_1}^{p\beta})^c(y_1)\cup X(y_1)]\cap[(\widetilde{SN}_{y_1}^{p\beta})^c(y_2)\cup X(y_2)]\}^c=(\widetilde{SN}_-^{p\beta}(X))^c(y_1)$.

$\widetilde{SN}_+^{p\beta}(X^c)(y_2)=(\widetilde{SN}_-^{p\beta}(X))^c(y_2)$ can be proved as above.

To sum up,  $\widetilde{SN}_+^{p\beta}(X^c)=(\widetilde{SN}_-^{p\beta}(X))^c$ hold for the case of $|U|=2$.
Other cases of $|U|<\infty$ can be proved by using the mathematical induction.

(3) By Theorem \ref{thm6} (6), for each $x\in U$, $\widetilde{SN}_-^{p\beta}(X\cap Y)(x)=\bigcap\limits_{y\in U}\{(\widetilde{SN}_x^{p\beta})^c(y)\cup (X\cap Y)(y)\}=_p\bigcap\limits_{y\in U}\{[(\widetilde{SN}_x^{p\beta})^c(y)\cup X(y)]\cap[(\widetilde{SN}_x^{p\beta})^c(y)\cup Y(y)]\}=_p\{\bigcap\limits_{y\in U}\{(\widetilde{SN}_x^{p\beta})^c(y)\cup X(y)\}\}\cap\{\bigcap\limits_{y\in U}\{(\widetilde{SN}_x^{p\beta})^c(y)\cup Y(y)\}\}=_p\widetilde{SN}_-^{p\beta}(X)(x)\cap\widetilde{SN}_-^{p\beta}(Y)(x)$.

By Theorem \ref{thm6} (6), for each $x\in U$, $\widetilde{SN}_+^{p\beta}(X\cup Y)(x)=\bigcup\limits_{y\in U}\{\widetilde{SN}_x^{p\beta}(y)\cap (X\cup Y)(y)\}=_p\bigcup\limits_{y\in U}\{[\widetilde{SN}_x^{p\beta}(y)\cap X(y)]\cup[\widetilde{SN}_x^{p\beta}(y)\cap Y(y)]\}=_p\{\bigcup\limits_{y\in U}\{\widetilde{SN}_x^{p\beta}(y)\cap X(y)\}\}\cup\{\bigcup\limits_{y\in U}\{\widetilde{SN}_x^{p\beta}(y)\cap Y(y)\}\}=_p\widetilde{SN}_+^{p\beta}(X)(x)\cup \widetilde{SN}_+^{p\beta}(Y)(x)$.

(4) For each $y\in U$, since $X\subset_p Y$, i.e., $X^+(y)\leqslant Y^+(y)$, then $((\widetilde{SN}_x^{p\beta})^c)^+(y)\vee X^+(y)\leqslant((\widetilde{SN}_x^{p\beta})^c)^+(y)\vee Y^+(y)$. It can deduce that $\min\{((\widetilde{SN}_x^{p\beta})^c\cup X)^+(y):y\in U\}\leqslant\min\{((\widetilde{SN}_x^{p\beta})^c\cup Y)^+(y):y\in U\}$, i.e., $\widetilde{SN}_-^{p\beta}(X)\subset_p\widetilde{SN}_-^{p\beta}(Y)$.

Since $X^+(y)\leqslant Y^+(y)$, then $(\widetilde{SN}_x^{p\beta})^+(y)\wedge X^+(y)\leqslant(\widetilde{SN}_x^{p\beta})^+(y)\wedge Y^+(y)$. It can deduce that $\max\{(\widetilde{SN}_x^{p\beta}\cap X)^+(y):y\in U\}\leqslant\max\{(\widetilde{SN}_x^{p\beta}\cap Y)^+(y):y\in U\}$, i.e., $\widetilde{SN}_+^{p\beta}(X)\subset_p\widetilde{SN}_+^{p\beta}(Y)$.

(5) $X\subset_p X\cup Y$ and $Y\subset_p X\cup Y$. By the results of (4), $\widetilde{SN}_-^{p\beta}(X)\subset_p\widetilde{SN}_-^{p\beta}(X\cup Y)$ and $\widetilde{SN}_-^{p\beta}(Y)\subset_p\widetilde{SN}_-^{p\beta}(X\cup Y)$. Then  $\widetilde{SN}_-^{p\beta}(X)\cup\widetilde{SN}_-^{p\beta}(Y)\subset_p\widetilde{SN}_-^{p\beta}(X\cup Y)$.

$X\cap Y\subset_p X$ and $X\cap Y\subset_p Y$. By the results of (4), $\widetilde{SN}_+^{p\beta}(X\cap Y)\subset_p\widetilde{SN}_+^{p\beta}(X)$ and $\widetilde{SN}_+^{p\beta}(X\cap Y)\subset_p\widetilde{SN}_+^{p\beta}(Y)$. Then $\widetilde{SN}_+^{p\beta}(X\cap Y)\subset_p\widetilde{SN}_+^{p\beta}(X)\cap\widetilde{SN}_+^{p\beta}(Y)$.

(6) Since $(\widetilde{SN}_x^{p\beta})^c(x)\subset_p X(x)$ for all $x\in U$, then $((\widetilde{SN}_x^{p\beta})^c)^+(x)\vee X^+(x)\leqslant X^+(x)\vee X^+(x)=X^+(x)$. Then,  $\min\{((\widetilde{SN}_x^{p\beta})^c)^+(y)\vee X^+(y):y\in U\}\leqslant((\widetilde{SN}_x^{p\beta})^c)^+(x)\vee X^+(x)\leqslant X^+(x)$, i.e., $\widetilde{SN}_-^{p\beta}(X)\subset_p X(x)$.

Since $X(x)\subset_p\widetilde{SN}_x^{p\beta}(x)$ for all $x\in U$, $X^+(x)=X^+(x)\wedge X^+(x)\leqslant X^+(x)\wedge(\widetilde{SN}_x^{p\beta})^+(x)$. Then, $X^+(x)\leqslant X^+(x)\wedge(\widetilde{SN}_x^{p\beta})^+(x)\leqslant\max\{(\widetilde{SN}_x^{p\beta})^+(y)\wedge X^+(y):y\in U\}$, i.e., $X(x)\subset_p\widetilde{SN}_+^{p\beta}(X)$.

Hence, $\widetilde{SN}_-^{p\beta}(X)\subset_p X(x)\subset_p\widetilde{SN}_+^{p\beta}(X)$.

(7) $(\widetilde{SN}_x^{p\beta})^c(x)\subset_p X(x)\subset_p\widetilde{SN}_x^{p\beta}(x)$ for all $x\in U$, by the results of (6), we have $\widetilde{SN}_-^{p\beta}(X)\subset_p X(x)\subset_p\widetilde{SN}_+^{p\beta}(X)$. By the results of (4), we have $\widetilde{SN}_-^{p\beta}(\widetilde{SN}_-^{p\beta}(X))\subset_p\widetilde{SN}_-^{p\beta}(X)\subset_p X\subset_p\widetilde{SN}_+^{p\beta}(X)\subset_p\widetilde{SN}_+^{p\beta}(\widetilde{SN}_+^{p\beta}(X))$.

(8) If $X\subset_p Y$, then $X\cup Y\subset_p Y\cup Y=_p Y$. By the results of (4), $\widetilde{SN}_-^{p\beta}(X\cup Y)\subset_p\widetilde{SN}_-^{p\beta}(Y)\subset_p\widetilde{SN}_-^{p\beta}(X)\cup\widetilde{SN}_-^{p\beta}(Y)$.

By the results of (5), $\widetilde{SN}_-^{p\beta}(X)\cup\widetilde{SN}_-^{p\beta}(Y)\subset_p\widetilde{SN}_-^{p\beta}(X\cup Y)$.

Then, $\widetilde{SN}_-^{p\beta}(X)\cup\widetilde{SN}_-^{p\beta}(Y)=_p\widetilde{SN}_-^{p\beta}(X\cup Y)$.

Since $X\subset_p Y$, then $X=_pX\cap X=_pX\cap Y\subset_pX\cap Y$. By the results of (4), $\widetilde{SN}_+^{p\beta}(X)\subset_p\widetilde{SN}_+^{p\beta}(X\cap Y)$. Then, we have $\widetilde{SN}_+^{p\beta}(X)\cap\widetilde{SN}_+^{p\beta}(Y)\subset_p\widetilde{SN}_+^{p\beta}(X)\subset_p\widetilde{SN}_+^{p\beta}(X\cap Y)$.

By the results of (5), $\widetilde{SN}_+^{p\beta}(X\cap Y)\subset_p\widetilde{SN}_+^{p\beta}(X)\cap\widetilde{SN}_+^{p\beta}(Y)$.

Hence, $\widetilde{SN}_+^{p\beta}(X\cap Y)=_p\widetilde{SN}_+^{p\beta}(X)\cap\widetilde{SN}_+^{p\beta}(Y)$.$\blacksquare$

\begin{pro}\label{proada} Let $\beta$ be a hesitant fuzzy number, $X,Y\in HF(U)$, $x,y\in U$,  the following statements hold in $T_a$.

(1)  $\widetilde{SN}_-^{a\beta}(H^U)=_nH^U$, $\widetilde{SN}_+^{a\beta}(H^{\emptyset})=_n H^{\emptyset}$.

(2) $\widetilde{SN}_-^{a\beta}(X^c)=(\widetilde{SN}_+^{a\beta}(X))^c$, $\widetilde{SN}_+^{a\beta}(X^c)=(\widetilde{SN}_-^{a\beta}(X))^c$.

(3)  $\widetilde{SN}_-^{a\beta}(X\cap Y)=_a\widetilde{SN}_-^{a\beta}(X)\cap \widetilde{SN}_-^{a\beta}(Y)$, $\widetilde{SN}_+^{a\beta}(X\cup Y)=_a\widetilde{SN}_+^{a\beta}(X)\cup \widetilde{SN}_+^{a\beta}(Y)$.

(4) If $X\subset_a Y$, then $\widetilde{SN}_-^{a\beta}(X)\subset_a\widetilde{SN}_-^{a\beta}(Y)$, $\widetilde{SN}_+^{a\beta}(X)\subset_a\widetilde{SN}_+^{a\beta}(Y)$.

(5) $\widetilde{SN}_-^{a\beta}(X)\cup\widetilde{SN}_-^{a\beta}(Y)\subset_a\widetilde{SN}_-^{a\beta}(X\cup Y)$, $\widetilde{SN}_+^{a\beta}(X\cap Y)\subset_a\widetilde{SN}_+^{a\beta}(X)\cap\widetilde{SN}_+^{a\beta}(Y)$.

(6) If $(\widetilde{SN}_x^{a\beta})^c(x)\subset_a X(x)\subset_a\widetilde{SN}_x^{a\beta}(x)$ for all $x\in U$, then $\widetilde{SN}_-^{a\beta}(X)\subset_a X(x)\subset_a\widetilde{SN}_+^{a\beta}(X)$.

(7) If $(\widetilde{SN}_x^{a\beta})^c(x)\subset_a X(x)\subset_a\widetilde{SN}_x^{a\beta}(x)$ for all $x\in U$, then $\widetilde{SN}_-^{a\beta}(\widetilde{SN}_-^{a\beta}(X))\subset_a\widetilde{SN}_-^{a\beta}(X)\subset_a X\subset_a\widetilde{SN}_+^{a\beta}(X)\subset_a\widetilde{SN}_+^{a\beta}(\widetilde{SN}_+^{a\beta}(X))$.

(8) If $X\subset_a Y$, $\widetilde{SN}_-^{a\beta}(X)\cup\widetilde{SN}_-^{a\beta}(Y)=_a\widetilde{SN}_-^{a\beta}(X\cup Y)$, $\widetilde{SN}_+^{a\beta}(X\cap Y)=_a\widetilde{SN}_+^{a\beta}(X)\cap\widetilde{SN}_+^{a\beta}(Y)$.
\end{pro}

{\bf\slshape Proof}  The proofs of (1) and (2) are similar to the proof of Proposition \ref{pro43} (1) and (2),  respectively.

(3) By Theorem \ref{thm6} (7), for each $x\in U$, $\widetilde{SN}_-^{a\beta}(X\cap Y)(x)=\bigcap\limits_{y\in U}\{(\widetilde{SN}_x^{a\beta})^c(y)\cup (X\cap Y)(y)\}=_a\bigcap\limits_{y\in U}\{[(\widetilde{SN}_x^{a\beta})^c(y)\cup X(y)]\cap[(\widetilde{SN}_x^{a\beta})^c(y)\cup Y(y)]\}=_a\{\bigcap\limits_{y\in U}\{(\widetilde{SN}_x^{a\beta})^c(y)\cup X(y)\}\}\cap\{\bigcap\limits_{y\in U}\{(\widetilde{SN}_x^{a\beta})^c(y)\cup Y(y)\}\}=_a\widetilde{SN}_-^{a\beta}(X)(x)\cap\widetilde{SN}_-^{a\beta}(Y)(x)$.

By Theorem \ref{thm6} (7), for each $x\in U$, $\widetilde{SN}_+^{a\beta}(X\cup Y)(x)=\bigcup\limits_{y\in U}\{\widetilde{SN}_x^{a\beta}(y)\cap (X\cup Y)(y)\}=_a\bigcup\limits_{y\in U}\{[\widetilde{SN}_x^{a\beta}(y)\cap X(y)]\cup[\widetilde{SN}_x^{a\beta}(y)\cap Y(y)]\}=_a\{\bigcup\limits_{y\in U}\{\widetilde{SN}_x^{a\beta}(y)\cap X(y)\}\}\cup\{\bigcup\limits_{y\in U}\{\widetilde{SN}_x^{a\beta}(y)\cap Y(y)\}\}=_a\widetilde{SN}_+^{a\beta}(X)(x)\cup \widetilde{SN}_+^{a\beta}(Y)(x)$.

(4)  For each $y\in U$, since $X\subset_a Y$, i.e., $X^-(y)\leqslant Y^-(y)$ and $X^+(y)\leqslant Y^+(y)$, then $((\widetilde{SN}_x^{a\beta})^c)^-(y)\vee X^-(y)\leqslant((\widetilde{SN}_x^{a\beta})^c)^-(y)\vee Y^-(y)$ and  $((\widetilde{SN}_x^{a\beta})^c)^+(y)\vee X^+(y)\leqslant((\widetilde{SN}_x^{a\beta})^c)^+(y)\vee Y^+(y)$. It can deduce that $\min\{((\widetilde{SN}_x^{a\beta})^c\cup X)^-(y):y\in U\}\leqslant\min\{((\widetilde{SN}_x^{a\beta})^c\cup Y)^-(y):y\in U\}$ and  $\min\{((\widetilde{SN}_x^{a\beta})^c\cup X)^+(y):y\in U\}\leqslant\min\{((\widetilde{SN}_x^{a\beta})^c\cup Y)^+(y):y\in U\}$, i.e., $\widetilde{SN}_-^{a\beta}(X)\subset_a\widetilde{SN}_-^{a\beta}(Y)$.

Since $X^-(y)\leqslant Y^-(y)$ and $X^+(y)\leqslant Y^+(y)$, then $(\widetilde{SN}_x^{a\beta})^-(y)\wedge X^-(y)\leqslant(\widetilde{SN}_x^{a\beta})^-(y)\wedge Y^-(y)$ and $(\widetilde{SN}_x^{a\beta})^+(y)\wedge X^+(y)\leqslant(\widetilde{SN}_x^{a\beta})^+(y)\wedge Y^+(y)$. It can deduce that $\max\{(\widetilde{SN}_x^{a\beta}\cap X)^-(y):y\in U\}\leqslant\max\{(\widetilde{SN}_x^{a\beta}\cap Y)^-(y):y\in U\}$ and $\max\{(\widetilde{SN}_x^{a\beta}\cap X)^+(y):y\in U\}\leqslant\max\{(\widetilde{SN}_x^{a\beta}\cap Y)^+(y):y\in U\}$, i.e., $\widetilde{SN}_+^{a\beta}(X)\subset_a\widetilde{SN}_+^{a\beta}(Y)$.

(5) $X\subset_a X\cup Y$ and $Y\subset_a X\cup Y$. By the results of (4), $\widetilde{SN}_-^{a\beta}(X)\subset_a\widetilde{SN}_-^{a\beta}(X\cup Y)$ and $\widetilde{SN}_-^{a\beta}(Y)\subset_a\widetilde{SN}_-^{a\beta}(X\cup Y)$. Then  $\widetilde{SN}_-^{a\beta}(X)\cup\widetilde{SN}_-^{a\beta}(Y)\subset_a\widetilde{SN}_-^{a\beta}(X\cup Y)$.

$X\cap Y\subset_a X$ and $X\cap Y\subset_a Y$. By the results of (4), $\widetilde{SN}_+^{a\beta}(X\cap Y)\subset_a\widetilde{SN}_+^{a\beta}(X)$ and $\widetilde{SN}_+^{a\beta}(X\cap Y)\subset_a\widetilde{SN}_+^{a\beta}(Y)$. Then $\widetilde{SN}_+^{a\beta}(X\cap Y)\subset_a\widetilde{SN}_+^{a\beta}(X)\cap\widetilde{SN}_+^{a\beta}(Y)$.

(6) Since $(\widetilde{SN}_x^{a\beta})^c(x)\subset_a X(x)$ for all $x\in U$, then $((\widetilde{SN}_x^{a\beta})^c)^-(x)\vee X^-(x)\leqslant X^-(x)\vee X^-(x)=X^-(x)$ and $((\widetilde{SN}_x^{a\beta})^c)^+(x)\vee X^+(x)\leqslant X^+(x)\vee X^+(x)=X^+(x)$. Then,  $\min\{((\widetilde{SN}_x^{a\beta})^c)^-(y)\vee X^-(y):y\in U\}\leqslant((\widetilde{SN}_x^{a\beta})^c)^-(x)\vee X^-(x)\leqslant X^-(x)$ and  $\min\{((\widetilde{SN}_x^{a\beta})^c)^+(y)\vee X^+(y):y\in U\}\leqslant((\widetilde{SN}_x^{a\beta})^c)^+(x)\vee X^+(x)\leqslant X^+(x)$, i.e., $\widetilde{SN}_-^{a\beta}(X)\subset_a X(x)$.

Since $X(x)\subset_a\widetilde{SN}_x^{a\beta}(x)$ for all $x\in U$, then $X^-(x)=X^-(x)\wedge X^-(x)\leqslant X^-(x)\wedge(\widetilde{SN}_x^{a\beta})^-(x)$ and $X^+(x)=X^+(x)\wedge X^+(x)\leqslant X^+(x)\wedge(\widetilde{SN}_x^{a\beta})^+(x)$.  It can deduce that $X^-(x)\leqslant X^-(x)\wedge(\widetilde{SN}_x^{a\beta})^-(x)\leqslant\max\{(\widetilde{SN}_x^{a\beta})^-(y)\wedge X^-(y):y\in U\}$ and  $X^+(x)\leqslant X^+(x)\wedge(\widetilde{SN}_x^{a\beta})^+(x)\leqslant\max\{(\widetilde{SN}_x^{a\beta})^+(y)\wedge X^+(y):y\in U\}$, i.e., $X(x)\subset_a\widetilde{SN}_+^{a\beta}(X)$.

Hence, $\widetilde{SN}_-^{a\beta}(X)\subset_a X(x)\subset_a\widetilde{SN}_+^{a\beta}(X)$.

(7) $(\widetilde{SN}_x^{a\beta})^c(x)\subset_a X(x)\subset_a\widetilde{SN}_x^{a\beta}(x)$ for all $x\in U$, by the results of (6), we have $\widetilde{SN}_-^{a\beta}(X)\subset_a X(x)\subset_a\widetilde{SN}_+^{a\beta}(X)$. By the results of (4), we have $\widetilde{SN}_-^{a\beta}(\widetilde{SN}_-^{a\beta}(X))\subset_a\widetilde{SN}_-^{a\beta}(X)\subset_a X\subset_a\widetilde{SN}_+^{a\beta}(X)\subset_a\widetilde{SN}_+^{a\beta}(\widetilde{SN}_+^{a\beta}(X))$.

(8) If $X\subset_a Y$, then $X\cup Y\subset_a Y\cup Y=_a Y$. By the results of (4), $\widetilde{SN}_-^{a\beta}(X\cup Y)\subset_a\widetilde{SN}_-^{a\beta}(Y)\subset_a\widetilde{SN}_-^{a\beta}(X)\cup\widetilde{SN}_-^{a\beta}(Y)$.

By the results of (5), $\widetilde{SN}_-^{a\beta}(X)\cup\widetilde{SN}_-^{a\beta}(Y)\subset_a\widetilde{SN}_-^{a\beta}(X\cup Y)$.

Then, $\widetilde{SN}_-^{a\beta}(X)\cup\widetilde{SN}_-^{a\beta}(Y)=_a\widetilde{SN}_-^{a\beta}(X\cup Y)$.

Since $X\subset_a Y$, then $X=_aX\cap X=_aX\cap Y\subset_aX\cap Y$. By the results of (4), $\widetilde{SN}_+^{a\beta}(X)\subset_a\widetilde{SN}_+^{a\beta}(X\cap Y)$. Then, we have $\widetilde{SN}_+^{a\beta}(X)\cap\widetilde{SN}_+^{a\beta}(Y)\subset_a\widetilde{SN}_+^{a\beta}(X)\subset_a\widetilde{SN}_+^{a\beta}(X\cap Y)$.

By the results of (5), $\widetilde{SN}_+^{a\beta}(X\cap Y)\subset_a\widetilde{SN}_+^{a\beta}(X)\cap\widetilde{SN}_+^{a\beta}(Y)$.

Hence, $\widetilde{SN}_+^{a\beta}(X\cap Y)=_a\widetilde{SN}_+^{a\beta}(X)\cap\widetilde{SN}_+^{a\beta}(Y)$.$\blacksquare$

\begin{pro} Let $\beta$ be a hesitant fuzzy number, $X,Y\in HF(U)$, $x,y\in U$,  the following statements hold in $T_m$.

(1)  $\widetilde{SN}_-^{m\beta}(H^U)=_nH^U$, $\widetilde{SN}_+^{m\beta}(H^{\emptyset})=_n H^{\emptyset}$.

(2)  $\widetilde{SN}_-^{m\beta}(X^c)=(\widetilde{SN}_+^{m\beta}(X))^c$,  $\widetilde{SN}_+^{m\beta}(X^c)=(\widetilde{SN}_-^{m\beta}(X))^c$.
\end{pro}

{\bf\slshape Proof}  The proofs of (1) and (2) are similar to the proofs of Proposition \ref{pro43} (1) and (2), respectively.$\blacksquare$

\begin{pro}\label{pro45} Let $\beta$ be a hesitant fuzzy number, $X,Y\in HF(U)$, $x,y\in U$,  the following statements hold in $T_s$.

(1)  $\widetilde{SN}_-^{s\beta}(H^U)=_nH^U$, $\widetilde{SN}_+^{s\beta}(H^{\emptyset})=_n H^{\emptyset}$.

(2)  $\widetilde{SN}_-^{s\beta}(X^c)=(\widetilde{SN}_+^{s\beta}(X))^c$, $\widetilde{SN}_+^{s\beta}(X^c)=(\widetilde{SN}_-^{s\beta}(X))^c$.

(3)  $\widetilde{SN}_-^{s\beta}(X\cap Y)=_a\widetilde{SN}_-^{s\beta}(X)\cap \widetilde{SN}_-^{s\beta}(Y)$, $\widetilde{SN}_+^{s\beta}(X\cup Y)=_a\widetilde{SN}_+^{s\beta}(X)\cup \widetilde{SN}_+^{s\beta}(Y)$.

(4) If $X\subset_s Y$, then $\widetilde{SN}_-^{s\beta}(X)\subset_a\widetilde{SN}_-^{s\beta}(Y)$, $\widetilde{SN}_+^{s\beta}(X)\subset_a\widetilde{SN}_+^{s\beta}(Y)$.

(5) $\widetilde{SN}_-^{s\beta}(X)\cup\widetilde{SN}_-^{s\beta}(Y)\subset_a\widetilde{SN}_-^{s\beta}(X\cup Y)$, $\widetilde{SN}_+^{s\beta}(X\cap Y)\subset_a\widetilde{SN}_+^{s\beta}(X)\cap\widetilde{SN}_+^{s\beta}(Y)$.

(6) If $(\widetilde{SN}_x^{s\beta})^c(x)\subset_s X(x)\subset_s\widetilde{SN}_x^{s\beta}(x)$ for all $x\in U$, then $\widetilde{SN}_-^{s\beta}(X)\subset_a X(x)\subset_a\widetilde{SN}_+^{s\beta}(X)$.

(7) If $(\widetilde{SN}_x^{s\beta})^c(x)\subset_s X(x)\subset_s\widetilde{SN}_x^{s\beta}(x)$ for all $x\in U$, then $\widetilde{SN}_-^{s\beta}(\widetilde{SN}_-^{s\beta}(X))\subset_a\widetilde{SN}_-^{s\beta}(X)\subset_a X\subset_a\widetilde{SN}_+^{s\beta}(X)\subset_a\widetilde{SN}_+^{s\beta}(\widetilde{SN}_+^{s\beta}(X))$.

(8) If $X\subset_s Y$, $\widetilde{SN}_-^{s\beta}(X)\cup\widetilde{SN}_-^{s\beta}(Y)=_a\widetilde{SN}_-^{s\beta}(X\cup Y)$, $\widetilde{SN}_+^{s\beta}(X\cap Y)=_a\widetilde{SN}_+^{s\beta}(X)\cap\widetilde{SN}_+^{s\beta}(Y)$.

{\bf\slshape Proof} Since $X\subset_s Y$ implies $X\subset_p Y$ and $X\subset_a Y$. The proofs of (1)-(2) are similar to the proofs of Proposition \ref{pro43} (1)-(2), respectively.  The proofs of (3)-(8) are similar to the proofs of  Proposition \ref{proada} (3)-(8),  respectively.$\blacksquare$
\end{pro}

\begin{pro}\label{pro46}  Let $\beta$ be a hesitant fuzzy number, $X,Y\in HF(U)$, $x,y\in U$,  the following statements hold in $T_t$.

(1)  $\widetilde{SN}_-^{t\beta}(H^U)=_nH^U$, $\widetilde{SN}_+^{t\beta}(H^{\emptyset})=_n H^{\emptyset}$.

(2) $\widetilde{SN}_-^{t\beta}(X^c)=(\widetilde{SN}_+^{t\beta}(X))^c$, $\widetilde{SN}_+^{t\beta}(X^c)=(\widetilde{SN}_-^{t\beta}(X))^c$.

(3)  $\widetilde{SN}_-^{t\beta}(X\cap Y)=_p\widetilde{SN}_-^{t\beta}(X)\cap \widetilde{SN}_-^{t\beta}(Y)$, $\widetilde{SN}_+^{t\beta}(X\cup Y)=_p\widetilde{SN}_+^{t\beta}(X)\cup \widetilde{SN}_+^{t\beta}(Y)$.

(4) If $X\subset_t Y$, then $\widetilde{SN}_-^{t\beta}(X)\subset_p\widetilde{SN}_-^{t\beta}(Y)$, $\widetilde{SN}_+^{t\beta}(X)\subset_p\widetilde{SN}_+^{t\beta}(Y)$.

(5) $\widetilde{SN}_-^{t\beta}(X)\cup\widetilde{SN}_-^{t\beta}(Y)\subset_p\widetilde{SN}_-^{t\beta}(X\cup Y)$, $\widetilde{SN}_+^{t\beta}(X\cap Y)\subset_p\widetilde{SN}_+^{t\beta}(X)\cap\widetilde{SN}_+^{t\beta}(Y)$.

(6) If $(\widetilde{SN}_x^{t\beta})^c(x)\subset_t X(x)\subset_t\widetilde{SN}_x^{t\beta}(x)$ for all $x\in U$, then $\widetilde{SN}_-^{t\beta}(X)\subset_p X(x)\subset_p\widetilde{SN}_+^{t\beta}(X)$.

(7) If $(\widetilde{SN}_x^{t\beta})^c(x)\subset_t X(x)\subset_t\widetilde{SN}_x^{t\beta}(x)$ for all $x\in U$, then $\widetilde{SN}_-^{t\beta}(\widetilde{SN}_-^{t\beta}(X))\subset_p\widetilde{SN}_-^{t\beta}(X)\subset_p X\subset_p\widetilde{SN}_+^{t\beta}(X)\subset_p\widetilde{SN}_+^{t\beta}(\widetilde{SN}_+^{t\beta}(X))$.

(8)  If $X\subset_t Y$, $\widetilde{SN}_-^{t\beta}(X)\cup\widetilde{SN}_-^{t\beta}(Y)=_p\widetilde{SN}_-^{t\beta}(X\cup Y)$, $\widetilde{SN}_+^{t\beta}(X\cap Y)=_p\widetilde{SN}_+^{t\beta}(X)\cap\widetilde{SN}_+^{t\beta}(Y)$.

{\bf\slshape Proof} Since $X\subset_t Y$ implies $X\subset_p Y$. The proofs of (1)-(8) are similar to the proofs of Proposition \ref{pro43} (1)-(8), respectively.$\blacksquare$
\end{pro}

\begin{pro}\label{pro47}  Let $\beta$ be a hesitant fuzzy number, $X,Y\in HF(U)$, $x,y\in U$,  the following statements hold in $T_n$.

(1)  $H^U\subset_n\widetilde{SN}_-^{n\beta}(H^U)$, $\widetilde{SN}_+^{n\beta}(H^{\emptyset})\subset_n H^{\emptyset}$.

(2)  $\widetilde{SN}_-^{n\beta}(X^c)=(\widetilde{SN}_+^{n\beta}(X))^c$, $\widetilde{SN}_+^{n\beta}(X^c)=(\widetilde{SN}_-^{n\beta}(X))^c$.

(3)  $\widetilde{SN}_-^{n\beta}(X\cap Y)=_a\widetilde{SN}_-^{n\beta}(X)\cap \widetilde{SN}_-^{n\beta}(Y)$, $\widetilde{SN}_+^{n\beta}(X\cup Y)=_a\widetilde{SN}_+^{n\beta}(X)\cup \widetilde{SN}_+^{n\beta}(Y)$.

(4) If $X\subset_n Y$, then $\widetilde{SN}_-^{n\beta}(X)\subset_a\widetilde{SN}_-^{n\beta}(Y)$, $\widetilde{SN}_+^{n\beta}(X)\subset_a\widetilde{SN}_+^{n\beta}(Y)$.

(5) $\widetilde{SN}_-^{n\beta}(X)\cup\widetilde{SN}_-^{n\beta}(Y)\subset_a\widetilde{SN}_-^{n\beta}(X\cup Y)$, $\widetilde{SN}_+^{n\beta}(X\cap Y)\subset_a\widetilde{SN}_+^{n\beta}(X)\cap\widetilde{SN}_+^{n\beta}(Y)$.

(6) If $(\widetilde{SN}_x^{n\beta})^c(x)\subset_n X(x)\subset_n\widetilde{SN}_x^{n\beta}(x)$ for all $x\in U$, then $\widetilde{SN}_-^{n\beta}(X)\subset_a X(x)\subset_a\widetilde{SN}_+^{n\beta}(X)$.

(7) If $(\widetilde{SN}_x^{n\beta})^c(x)\subset_n X(x)\subset_n\widetilde{SN}_x^{n\beta}(x)$ for all $x\in U$, then $\widetilde{SN}_-^{n\beta}(\widetilde{SN}_-^{n\beta}(X))\subset_a\widetilde{SN}_-^{n\beta}(X)\subset_a X\subset_a\widetilde{SN}_+^{n\beta}(X)\subset_a\widetilde{SN}_+^{n\beta}(\widetilde{SN}_+^{n\beta}(X))$.

(8) If $X\subset_n Y$, $\widetilde{SN}_-^{n\beta}(X)\cup\widetilde{SN}_-^{n\beta}(Y)=_a\widetilde{SN}_-^{n\beta}(X\cup Y)$, $\widetilde{SN}_+^{n\beta}(X\cap Y)=_a\widetilde{SN}_+^{n\beta}(X)\cap\widetilde{SN}_+^{n\beta}(Y)$.

{\bf\slshape Proof}  Since $X\subset_n Y$ implies $X\subset_p Y$ and $X\subset_a Y$. The proofs of (1)-(2) are similar to the proofs of Proposition \ref{pro43} (1)-(2), respectively.  The proofs of (3)-(8) are similar to the proofs of  Proposition \ref{proada} (3)-(8),  respectively.$\blacksquare$
\end{pro}

Although the conditions $X\subset_s Y$ and $X\subset_n Y$ are stronger than the condition $X\subset_a Y$, Propositions \ref{pro45} and \ref{pro47} do not obtain stronger conclusions than the conclusions of Proposition \ref{proada}, because the distributive law of hesitant fuzzy sets is satisfied only under the $=_p$ and $=_a$ relationships (shown as Theorem \ref{thm6} (6) and (7)).

\begin{pro}\label{pro48}   Let $\beta_1$ and $\beta_2$ be two hesitant fuzzy numbers, $X\in HF(U)$, $x,y\in U$, the the following statements hold.

(1) If $\beta_1\subset_p\beta_2$, $A\in T^1_p=(U,F,A)_{\beta_1}$ and $A\in T^2_p=(U,F,A)_{\beta_2}$, then $\widetilde{SN}_+^{p\beta_1}(X)\subset_a\widetilde{SN}_+^{p\beta_2}(X)$ and $\widetilde{SN}_-^{p\beta_2}(X)\subset_a\widetilde{SN}_-^{p\beta_1}(X)$.

(2) If $\beta_1\subset_a\beta_2$, $A\in T^1_a=(U,F,A)_{\beta_1}$ and $A\in T^2_a=(U,F,A)_{\beta_2}$, then $\widetilde{SN}_+^{a\beta_1}(X)\subset_a\widetilde{SN}_+^{a\beta_2}(X)$ and $\widetilde{SN}_-^{a\beta_2}(X)\subset_a\widetilde{SN}_-^{a\beta_1}(X)$.

(3) If $\beta_1\subset_m\beta_2$, $A\in T^1_m=(U,F,A)_{\beta_1}$ and $A\in T^2_m=(U,F,A)_{\beta_2}$, then $\widetilde{SN}_+^{m\beta_1}(X)\subset_a\widetilde{SN}_+^{m\beta_2}(X)$ and $\widetilde{SN}_-^{m\beta_2}(X)\subset_a\widetilde{SN}_-^{m\beta_1}(X)$.

(4) If $\beta_1\subset_s\beta_2$, $A\in T^1_s=(U,F,A)_{\beta_1}$ and $A\in T^2_s=(U,F,A)_{\beta_2}$, then $\widetilde{SN}_+^{s\beta_1}(X)\subset_a\widetilde{SN}_+^{s\beta_2}(X)$ and $\widetilde{SN}_-^{s\beta_2}(X)\subset_a\widetilde{SN}_-^{s\beta_1}(X)$.

(5) If $\beta_1\subset_t\beta_2$, $A\in T^1_t=(U,F,A)_{\beta_1}$ and $A\in T^2_t=(U,F,A)_{\beta_2}$, then $\widetilde{SN}_+^{t\beta_1}(X)\subset_a\widetilde{SN}_+^{t\beta_2}(X)$ and $\widetilde{SN}_-^{t\beta_2}(X)\subset_a\widetilde{SN}_-^{t\beta_1}(X)$.

(6) If $\beta_1\subset_n\beta_2$, $A\in T^1_n=(U,F,A)_{\beta_1}$ and $A\in T^2_n=(U,F,A)_{\beta_2}$, then $\widetilde{SN}_+^{n\beta_1}(X)\subset_a\widetilde{SN}_+^{n\beta_2}(X)$ and $\widetilde{SN}_-^{n\beta_2}(X)\subset_a\widetilde{SN}_-^{n\beta_1}(X)$.

{\bf\slshape Proof} (1) Since $\beta_1\subset_p\beta_2$, by Proposition \ref{prop2} (1), we have $\widetilde{SN}^{p\beta_1}_x\subset_a\widetilde{SN}^{p\beta_2}_x$, then $(\widetilde{SN}^{p\beta_2}_x)^c(y)\subset_a(\widetilde{SN}^{p\beta_1}_x)^c(y)$.

Based on the results above, $\widetilde{SN}^{p\beta_1}_x(y)\cap X(y)\subset_a\widetilde{SN}^{p\beta_2}_x(y)\cap X(y)$ and $(\widetilde{SN}^{p\beta_2}_x)^c(y)\cup X(y)\subset_a(\widetilde{SN}^{p\beta_1}_x)^c(y)\cup X(y)$, then $\widetilde{SN}_+^{p\beta_1}(X)\subset_a\widetilde{SN}_+^{p\beta_2}(X)$ and $\widetilde{SN}_-^{p\beta_2}(X)\subset_a\widetilde{SN}_-^{p\beta_1}(X)$.

(2)-(6) can be proved by the similar method of (1) with using the results of Proposition \ref{prop2} (2)-(6).$\blacksquare$
\end{pro}

\begin{defn}  Let $\beta$ be a hesitant fuzzy number and $X\sqsubset U$ be an object set, $x,y\in U$.

(1) $\overline{SN}_-^{p\beta}(X)=\{x\in U:\overline{SN}^{p\beta}_x\sqsubset X\}$ and $\overline{SN}_+^{p\beta}(X)=\{x\in U:\overline{SN}^{p\beta}_x\sqcap X\neq\emptyset\}$ are the $p$-lower approximation and $p$-upper approximation of the object set $X$ in $T_p=(U,F,A)_{\beta}$, respectively.

(2) $\overline{SN}_-^{a\beta}(X)=\{x\in U:\overline{SN}^{a\beta}_x\sqsubset X\}$ and $\overline{SN}_+^{a\beta}(X)=\{x\in U:\overline{SN}^{a\beta}_x\sqcap X\neq\emptyset\}$ are the $a$-lower approximation and $a$-upper approximation of the object set $X$ in $T_a=(U,F,A)_{\beta}$, respectively.

(3) $\overline{SN}_-^{m\beta}(X)=\{x\in U:\overline{SN}^{m\beta}_x\sqsubset X\}$ and $\overline{SN}_+^{m\beta}(X)=\{x\in U:\overline{SN}^{m\beta}_x\sqcap X\neq\emptyset\}$ are the $m$-lower approximation and $m$-upper approximation of the object set $X$ in $T_m=(U,F,A)_{\beta}$, respectively.

(4) $\overline{SN}_-^{s\beta}(X)=\{x\in U:\overline{SN}^{s\beta}_x\sqsubset X\}$ and $\overline{SN}_+^{s\beta}(X)=\{x\in U:\overline{SN}^{s\beta}_x\sqcap X\neq\emptyset\}$ are the $s$-lower approximation and $s$-upper approximation of the object set $X$ in $T_s=(U,F,A)_{\beta}$, respectively.

(5) $\overline{SN}_-^{t\beta}(X)=\{x\in U:\overline{SN}^{t\beta}_x\sqsubset X\}$ and $\overline{SN}_+^{t\beta}(X)=\{x\in U:\overline{SN}^{t\beta}_x\sqcap X\neq\emptyset\}$ are the $t$-lower approximation and $t$-upper approximation of the object set $X$ in $T_t=(U,F,A)_{\beta}$, respectively.

(6) $\overline{SN}_-^{n\beta}(X)=\{x\in U:\overline{SN}^{n\beta}_x\sqsubset X\}$ and $\overline{SN}_+^{n\beta}(X)=\{x\in U:\overline{SN}^{n\beta}_x\sqcap X\neq\emptyset\}$ are the $n$-lower approximation and $n$-upper approximation of the object set $X$ in $T_n=(U,F,A)_{\beta}$, respectively.
\end{defn}

\begin{exam}\label{exam410} The triple $(U,F,E)$ are shown in Table 2. Let $\beta=\{0.5,0.4,0.3\}$.

We take the results in Example \ref{examp1} and Example \ref{exam319}   to analyse the following cases.

(1) Let $A=\{e_1,e_2\}$, $T_p=(U,F,A)_{\beta}$ is a a hesitant fuzzy soft $p\beta$-covering approximation space. Let $X=\{x_1,x_2\}$.

Since $\beta\subset_pF(e_1)(x_1)$ and $\beta\subset_pF(e_2)(x_1)$, then $\widetilde{SN}_{x_1}^{p\beta}=F(e_1)\cap F(e_2)$.  $\beta\subset_p\widetilde{SN}_{x_1}^{p\beta}(y)$ hold for $y\in \{x_1,x_3,x_4,x_5\}=\overline{SN}^{p\beta}_{x_1}\not\sqsubset X$.

Since $\beta\subset_pF(e_1)(x_2)$ and $\beta\not\subset_pF(e_2)(x_2)$, then $\widetilde{SN}_{x_2}^{p\beta}=F(e_1)$.  $\beta\subset_p\widetilde{SN}_{x_2}^{p\beta}(y)$ hold for $y\in U=\overline{SN}^{p\beta}_{x_2}\not\sqsubset X$.

For $y\in\{x_3,x_4,x_5\}$, by Proposition \ref{pro320} (1), $y\in\overline{SN}^{p\beta}_y$ and $y\not\in X$. It can obtain that $\overline{SN}^{p\beta}_{x_3}\not\sqsubset X$, $\overline{SN}^{p\beta}_{x_4}\not\sqsubset X$ and $\overline{SN}^{p\beta}_{x_5}\not\sqsubset X$.

To sum up, $\overline{SN}_-^{p\beta}(X)=\emptyset$.

$x_1\in\overline{SN}^{p\beta}_{x}$ for all $x\in U$, then $\overline{SN}^{p\beta}_{x}\sqcap X\neq\emptyset$ for all $x\in U$, i.e., $\overline{SN}_+^{p\beta}(X)=U$.

(2) Let $A=\{e_1,e_2\}$, $T_a=(U,F,A)_{\beta}$ is a hesitant fuzzy soft $a\beta$-covering approximation space. Let $X=\{x_1,x_2\}$.

$\widetilde{SN}_{x_1}^{a\beta}=F(e_1)\cap F(e_2)$,  $\overline{SN}^{a\beta}_{x_1}=\{x_1\}\sqsubset X$. $\overline{SN}^{a\beta}_{x}\not\sqsubset X$ for all $x\in U-\{x_1\}$, then $\overline{SN}_-^{a\beta}(X)=\{x_1\}$.

$x_1\in\overline{SN}^{a\beta}_{x}$ for all $x\in U$, then $\overline{SN}^{a\beta}_{x}\sqcap X\neq\emptyset$ for all $x\in U$, i.e., $\overline{SN}_+^{a\beta}(X)=U$.

(3) Let $A=\{e_3,e_4\}$, $\beta\subset_m\bigcup\limits_{e\in A}F(e)$, then $T_m=(U,F,A)_{\beta}$ is a hesitant fuzzy soft $m\beta$-covering approximation space.

Let $X=\{x_1,x_2,x_4,x_5\}$. Since $\beta\subset_m F(e_3)(x_3)$ and $\beta\subset_m F(e_4)(x_3)$, then $\widetilde{SN}_{x_3}^{m\beta}=F(e_3)\cap F(e_4)=\frac{0.7,0.7,0.5,0.2}{x_1}+\frac{0.5,0.4,0.1}{x_2}+\frac{0.5,0.2}{x_3}+\frac{0.5,0.5,0.4}{x_4}+\frac{0.6,0.6,0.6,0.5,0.2}{x_5}$. Since
$\beta\subset_m\widetilde{SN}_{x_3}^{m\beta}(y)$ for $y\in \{x_1,x_4,x_5\}$, then $\overline{SN}^{m\beta}_{x_3}=\{x_1,x_4,x_5\}\sqsubset X$. It is said that $x_3\in\overline{SN}_-^{m\beta}(X)$. However, $x_3\not\in X$, then $\overline{SN}_-^{m\beta}(X)\not\sqsubset X$.

Let $Y=\{x_3\}$. $x_3\not\in\overline{SN}^{m\beta}_{x_3}$, $\overline{SN}^{m\beta}_{x_3}\sqcap Y=\emptyset$, then $x_3\not\in\overline{SN}_+^{m\beta}(Y)$, then $Y\not\sqsubset\overline{SN}_+^{m\beta}(Y)$.

(4) Let $A=\{e_4,e_9\}$, $\beta\subset_s\bigcup\limits_{e\in A}F(e)$, then $T_s=(U,F,A)_{\beta}$ is  a hesitant fuzzy soft $s\beta$-covering approximation space.

Let $X=\{x_1,x_2,x_3,x_4\}$. Since $\beta\subset_s F(e_4)(x_5)$ and $\beta\subset_s F(e_9)(x_5)$, then $\widetilde{SN}_{x_5}^{s\beta}=F(e_4)\cap F(e_9)$. $\overline{SN}^{s\beta}_{x_5}=\{x_4\}\sqsubset X$, then $x_5\in\overline{SN}_-^{s\beta}(X)$. It shows that $\overline{SN}_-^{s\beta}(X)\not\sqsubset X$.

Let $Y=\{x_5\}$. $x_5\not\in\overline{SN}^{s\beta}_{x_5}$, $\overline{SN}^{s\beta}_{x_5}\sqcap Y=\emptyset$, then $x_5\not\in\overline{SN}_+^{s\beta}(Y)$. It shows that $Y\not\sqsubset\overline{SN}_+^{s\beta}(Y)$.

(5) Let $A=\{e_6,e_8\}$, $\beta\subset_t\bigcup\limits_{e\in A}F(e)$. $T_t=(U,F,A)_{\beta}$ is a hesitant fuzzy soft $t\beta$-covering approximation space. Let $X=\{x_1,x_2\}$.

For all $x\in U$, $\beta\not\subset_t F(e_6)(x)$ and $\beta\subset_t F(e_8)(x)$, then $\widetilde{SN}^{t\beta}_{x}=F(e_8)$.

Since $\beta\subset_t \widetilde{SN}^{t\beta}_{x}(y)=F(e_8)(y)$ for all $x,y\in U$, then $\overline{SN}^{t\beta}_{x}=U$. It is easy to obtain that $\overline{SN}_-^{t\beta}(X)=\emptyset$ and $\overline{SN}_+^{t\beta}(X)=U$.

(6) Let $A=\{e_6,e_7\}$, $\beta\subset_n\bigcup\limits_{e\in A}F(e)$. $T_n=(U,F,A)_{\beta}$ is a hesitant fuzzy soft $n\beta$-covering approximation space.  Let $X=\{x_1,x_2\}$.

For all $x\in U$, $\beta\subset_n F(e_6)(x)$ and $\beta\not\subset_n F(e_7)(x)$, then $\widetilde{SN}^{n\beta}_{x}=F(e_6)$.

Since $\beta\subset_n \widetilde{SN}^{n\beta}_{x}(y)=F(e_6)(y)$ for all $x,y\in U$, then $\overline{SN}^{n\beta}_{x}=U$. It is easy to obtain that $\overline{SN}_-^{n\beta}(X)=\emptyset$ and $\overline{SN}_+^{n\beta}(X)=U$.

\end{exam}

\begin{pro}\label{pro411} Let $\beta$ be a hesitant fuzzy number and $X,Y\sqsubset U$, the the following statements hold in $T_p$ for $x,y\in U$.

(1) $\overline{SN}_-^{p\beta}(\emptyset)=\emptyset$, $\overline{SN}_-^{p\beta}(U)=U$.

(2) $\overline{SN}_+^{p\beta}(\emptyset)=\emptyset$, $\overline{SN}_+^{p\beta}(U)=U$.

(3) If $X\sqsubset Y$, then $\overline{SN}_-^{p\beta}(X)\sqsubset\overline{SN}_-^{p\beta}(Y)$ and $\overline{SN}_+^{p\beta}(X)\sqsubset\overline{SN}_+^{p\beta}(Y)$

(4) $\overline{SN}_-^{p\beta}(X)\sqcup\overline{SN}_-^{p\beta}(Y)\sqsubset\overline{SN}_-^{p\beta}(X\sqcup Y)$, $\overline{SN}_+^{p\beta}(X\sqcap Y)\sqsubset\overline{SN}_+^{p\beta}(X)\sqcap\overline{SN}_+^{p\beta}(Y)$.

(5) $\overline{SN}_-^{p\beta}(X)\sqcap\overline{SN}_-^{p\beta}(Y)=\overline{SN}_-^{p\beta}(X\sqcap Y)$, $\overline{SN}_+^{p\beta}(X\sqcup Y)=\overline{SN}_+^{p\beta}(X)\sqcup\overline{SN}_+^{p\beta}(Y)$.

(6) $\overline{SN}_-^{p\beta}(X^c)=(\overline{SN}_+^{p\beta}(X))^c$, $\overline{SN}_+^{p\beta}(X^c)=(\overline{SN}_-^{p\beta}(X))^c$.

(7) $\overline{SN}_-^{p\beta}(X)\sqsubset X\sqsubset\overline{SN}_+^{p\beta}(X)$.
\end{pro}

{\bf\slshape Proof} (1) and (2) are obvious.

(3) If $x\in\overline{SN}_-^{p\beta}(X)$, then $\overline{SN}^{p\beta}_x\sqsubset X\sqsubset Y$, i.e., $x\in\overline{SN}_-^{p\beta}(Y)$. Then $\overline{SN}_-^{p\beta}(X)\sqsubset \overline{SN}_-^{p\beta}(Y)$.

If $x\in\overline{SN}_+^{p\beta}(X)$, then $\emptyset\neq\overline{SN}^{p\beta}_x\sqcap X\sqsubset \overline{SN}^{p\beta}_x\sqcap Y$, i.e., $x\in\overline{SN}_+^{p\beta}(Y)$. Then $\overline{SN}_+^{p\beta}(X)\sqsubset \overline{SN}_+^{p\beta}(Y)$.

(4) Since $X\sqsubset X\sqcup Y$ and $Y\sqsubset X\sqcup Y$, by the results of (3), $\overline{SN}_-^{p\beta}(X)\sqsubset\overline{SN}_-^{p\beta}(X\sqcup Y)$ and $\overline{SN}_-^{p\beta}(Y)\sqsubset\overline{SN}_-^{p\beta}(X\sqcup Y)$. Then  $\overline{SN}_-^{p\beta}(X)\sqcup\overline{SN}_-^{p\beta}(Y)\sqsubset\overline{SN}_-^{p\beta}(X\sqcup Y)$.

Since $X\sqcap Y\sqsubset X$ and $X\sqcap Y\sqsubset Y$, by the results of (3), $\overline{SN}_+^{p\beta}(X\sqcap Y)\sqsubset\overline{SN}_+^{p\beta}(X)$ and $\overline{SN}_+^{p\beta}(X\sqcap Y)\sqsubset\overline{SN}_+^{p\beta}(Y)$. Then $\overline{SN}_+^{p\beta}(X\sqcap Y)\sqsubset\overline{SN}_+^{p\beta}(X)\sqcap\overline{SN}_+^{p\beta}(Y)$.

(5) If $x\in\overline{SN}_-^{p\beta}(X)\sqcap\overline{SN}_-^{p\beta}(Y)$, i.e., $x\in\overline{SN}_-^{p\beta}(X)$ and $x\in\overline{SN}_-^{p\beta}(Y)$, then $\overline{SN}^{p\beta}_x\sqsubset X$ and $\overline{SN}^{p\beta}_x\sqsubset Y$. It means that $\overline{SN}^{p\beta}_x\sqsubset X\sqcap Y$, i.e., $x\in\overline{SN}_-^{p\beta}(X\sqcap Y)$. Then $\overline{SN}_-^{p\beta}(X)\sqcap\overline{SN}_-^{p\beta}(Y)\sqsubset\overline{SN}_-^{p\beta}(X\sqcap Y)$.

On the other hand, $X\sqcap Y\sqsubset X$ and $X\sqcap Y\sqsubset Y$, by the results of (3), $\overline{SN}_-^{p\beta}(X\sqcap Y)\sqsubset\overline{SN}_-^{p\beta}(X)\sqcap\overline{SN}_-^{p\beta}(Y)$.

Hence, $\overline{SN}_-^{p\beta}(X\sqcap Y)=\overline{SN}_-^{p\beta}(X)\sqcap\overline{SN}_-^{p\beta}(Y)$.

If $x\in\overline{SN}_+^{p\beta}(X\sqcup Y)$, i.e., $\overline{SN}^{p\beta}_x\sqcap(X\sqcup Y)\neq\emptyset$, then there is $y\in U$ such that $y\in\overline{SN}^{p\beta}_x$ and $y\in X\sqcup Y$, it means that at least one of $y\in X$ and $y\in Y$ holds, then we can obtain that at least one of $\overline{SN}^{p\beta}_x\sqcap X\neq\emptyset$ and $\overline{SN}^{p\beta}_x\sqcap Y\neq\emptyset$ holds, i.e., $x\in\overline{SN}_+^{p\beta}(X)\sqcup\overline{SN}_+^{p\beta}(Y)$. Then $\overline{SN}_+^{p\beta}(X\sqcup Y)\sqsubset\overline{SN}_+^{p\beta}(X)\sqcup\overline{SN}_+^{p\beta}(Y)$.

On the other hand, $X\sqsubset X\sqcup Y$ and $Y\sqsubset X\sqcup Y$, by the results of (3), $\overline{SN}_+^{p\beta}(X)\sqcup\overline{SN}_+^{p\beta}(Y)\sqsubset\overline{SN}_+^{p\beta}(X\sqcup Y)$.

Hence, $\overline{SN}_+^{p\beta}(X)\sqcup\overline{SN}_+^{p\beta}(Y)=\overline{SN}_+^{p\beta}(X\sqcup Y)$.

(6) If $x\in\overline{SN}_-^{p\beta}(X^c)$, i.e., $\overline{SN}^{p\beta}_x\sqsubset X^c$, then $\overline{SN}^{p\beta}_x\sqcap X=\emptyset$, i.e., $x\not\in\overline{SN}_+^{p\beta}(X)$, thus $x\in(\overline{SN}_+^{p\beta}(X))^c$. Then  $\overline{SN}_-^{p\beta}(X^c)\sqsubset (\overline{SN}_+^{p\beta}(X))^c$.

On the other hand, $x\in(\overline{SN}_+^{p\beta}(X))^c$, i.e., $x\not\in\overline{SN}_+^{p\beta}(X)$, thus $\overline{SN}^{p\beta}_x\sqcap X=\emptyset$. Then $\overline{SN}^{p\beta}_x\sqsubset X^c$, i.e., $x\in\overline{SN}_-^{p\beta}(X^c)$. Then  $(\overline{SN}_+^{p\beta}(X))^c\sqsubset\overline{SN}_-^{p\beta}(X^c)$.

Hence, $\overline{SN}_-^{p\beta}(X^c)=(\overline{SN}_+^{p\beta}(X))^c$.

If  $x\in\overline{SN}_+^{p\beta}(X^c)$, i.e., $\overline{SN}^{p\beta}_x\sqcap X^c\neq\emptyset$, then $\overline{SN}^{p\beta}_x\not\sqsubset X$, i.e., $x\not\in\overline{SN}_-^{p\beta}(X)$ and $x\in(\overline{SN}_-^{p\beta}(X))^c$. Then $\overline{SN}_+^{p\beta}(X^c)\sqsubset(\overline{SN}_-^{p\beta}(X))^c$.

On the other hand, if $x\in(\overline{SN}_-^{p\beta}(X))^c$, i.e., $x\not\in\overline{SN}_-^{p\beta}(X)$ and $\overline{SN}^{p\beta}_x\not\sqsubset X$. Thus  $\overline{SN}^{p\beta}_x\sqcap X^c\neq\emptyset$, i.e., $x\in\overline{SN}_+^{p\beta}(X^c)$. Then $(\overline{SN}_-^{p\beta}(X))^c\sqsubset\overline{SN}_+^{p\beta}(X^c)$.

Hence, $\overline{SN}_+^{p\beta}(X^c)=(\overline{SN}_-^{p\beta}(X))^c$.

(7) By Proposition \ref{pro320} (1), $x\in\overline{SN}^{p\beta}_x$ for all $x\in U$.

For a random $y\not\in X$ ($y\in X^c$), since $y\in\overline{SN}^{p\beta}_y$, then $\overline{SN}^{p\beta}_y\not\sqsubset X$, i.e., $y\not\in\overline{SN}_-^{p\beta}(X)$ ($y\in(\overline{SN}_-^{p\beta}(X))^c$), then $X^c\sqsubset(\overline{SN}_-^{p\beta}(X))^c$, i.e.,  $\overline{SN}_-^{p\beta}(X)\sqsubset X$.

For all $x\in X$, since $x\in\overline{SN}^{p\beta}_x$, then $\emptyset\neq\{x\}\sqsubset\overline{SN}^{p\beta}_x\sqcap X$, i.e., $x\in\overline{SN}_+^{p\beta}(X)$. Then $X\sqsubset\overline{SN}_+^{p\beta}(X)$.

To sum up, $\overline{SN}_-^{p\beta}(X)\sqsubset X\sqsubset\overline{SN}_+^{p\beta}(X)$.$\blacksquare$

\begin{pro} Let $\beta$ be a hesitant fuzzy number and $X,Y\sqsubset U$, the the following statements hold in $T_a$ for $x,y\in U$.

(1) $\overline{SN}_-^{a\beta}(\emptyset)=\emptyset$, $\overline{SN}_-^{a\beta}(U)=U$.

(2) $\overline{SN}_+^{a\beta}(\emptyset)=\emptyset$, $\overline{SN}_+^{a\beta}(U)=U$.

(3) If $X\sqsubset Y$, then $\overline{SN}_-^{a\beta}(X)\sqsubset\overline{SN}_-^{a\beta}(Y)$ and $\overline{SN}_+^{a\beta}(X)\sqsubset\overline{SN}_+^{a\beta}(Y)$

(4) $\overline{SN}_-^{a\beta}(X)\sqcup\overline{SN}_-^{a\beta}(Y)\sqsubset\overline{SN}_-^{a\beta}(X\sqcup Y)$, $\overline{SN}_+^{a\beta}(X\sqcap Y)\sqsubset\overline{SN}_+^{a\beta}(X)\sqcap\overline{SN}_+^{a\beta}(Y)$.

(5) $\overline{SN}_-^{a\beta}(X)\sqcap\overline{SN}_-^{a\beta}(Y)=\overline{SN}_-^{a\beta}(X\sqcap Y)$, $\overline{SN}_+^{a\beta}(X\sqcup Y)=\overline{SN}_+^{a\beta}(X)\sqcup\overline{SN}_+^{a\beta}(Y)$.

(6) $\overline{SN}_-^{a\beta}(X^c)=(\overline{SN}_+^{a\beta}(X))^c$, $\overline{SN}_+^{a\beta}(X^c)=(\overline{SN}_-^{a\beta}(X))^c$.

(7) $\overline{SN}_-^{a\beta}(X)\sqsubset X\sqsubset\overline{SN}_+^{a\beta}(X)$.

{\bf\slshape Proof} The proofs of (1)-(7) are similar to the proofs of Proposition \ref{pro411} (1)-(7).$\blacksquare$
\end{pro}

\begin{pro} Let $\beta$ be a hesitant fuzzy number and $X,Y\sqsubset U$, the the following statements hold in $T_m$ for $x,y\in U$.

(1) $\overline{SN}_-^{m\beta}(\emptyset)=\emptyset$, $\overline{SN}_-^{m\beta}(U)=U$.

(2) $\overline{SN}_+^{m\beta}(\emptyset)=\emptyset$, $\overline{SN}_+^{m\beta}(U)=U$.

(3) If $X\sqsubset Y$, then $\overline{SN}_-^{m\beta}(X)\sqsubset\overline{SN}_-^{m\beta}(Y)$ and $\overline{SN}_+^{m\beta}(X)\sqsubset\overline{SN}_+^{m\beta}(Y)$

(4) $\overline{SN}_-^{m\beta}(X)\sqcup\overline{SN}_-^{m\beta}(Y)\sqsubset\overline{SN}_-^{m\beta}(X\sqcup Y)$, $\overline{SN}_+^{m\beta}(X\sqcap Y)\sqsubset\overline{SN}_+^{m\beta}(X)\sqcap\overline{SN}_+^{m\beta}(Y)$.

(5) $\overline{SN}_-^{m\beta}(X)\sqcap\overline{SN}_-^{m\beta}(Y)=\overline{SN}_-^{m\beta}(X\sqcap Y)$, $\overline{SN}_+^{m\beta}(X\sqcup Y)=\overline{SN}_+^{m\beta}(X)\sqcup\overline{SN}_+^{m\beta}(Y)$.

(6) $\overline{SN}_-^{m\beta}(X^c)=(\overline{SN}_+^{m\beta}(X))^c$, $\overline{SN}_+^{m\beta}(X^c)=(\overline{SN}_-^{m\beta}(X))^c$.
\end{pro}

{\bf\slshape Proof} The proofs of (1)-(6) are similar to the proofs of Proposition \ref{pro411} (1)-(6).$\blacksquare$

\begin{pro}  Let $\beta$ be a hesitant fuzzy number and $X,Y\sqsubset U$, the the following statements hold in $T_s$ for $x,y\in U$.

(1) $\overline{SN}_-^{s\beta}(\emptyset)=\emptyset$, $\overline{SN}_-^{s\beta}(U)=U$.

(2) $\overline{SN}_+^{s\beta}(\emptyset)=\emptyset$, $\overline{SN}_+^{s\beta}(U)=U$.

(3) If $X\sqsubset Y$, then $\overline{SN}_-^{s\beta}(X)\sqsubset\overline{SN}_-^{s\beta}(Y)$ and $\overline{SN}_+^{s\beta}(X)\sqsubset\overline{SN}_+^{s\beta}(Y)$

(4) $\overline{SN}_-^{s\beta}(X)\sqcup\overline{SN}_-^{s\beta}(Y)\sqsubset\overline{SN}_-^{s\beta}(X\sqcup Y)$, $\overline{SN}_+^{s\beta}(X\sqcap Y)\sqsubset\overline{SN}_+^{s\beta}(X)\sqcap\overline{SN}_+^{s\beta}(Y)$.

(5) $\overline{SN}_-^{s\beta}(X)\sqcap\overline{SN}_-^{s\beta}(Y)=\overline{SN}_-^{s\beta}(X\sqcap Y)$, $\overline{SN}_+^{s\beta}(X\sqcup Y)=\overline{SN}_+^{s\beta}(X)\sqcup\overline{SN}_+^{s\beta}(Y)$.

(6) $\overline{SN}_-^{s\beta}(X^c)=(\overline{SN}_+^{s\beta}(X))^c$, $\overline{SN}_+^{s\beta}(X^c)=(\overline{SN}_-^{s\beta}(X))^c$.
\end{pro}

{\bf\slshape Proof} The proofs of (1)-(6) are similar to the proofs of Proposition \ref{pro411} (1)-(6).$\blacksquare$

\begin{pro} Let $\beta$ be a hesitant fuzzy number and $X,Y\sqsubset U$, the the following statements hold in $T_t$ for $x,y\in U$.

(1) $\overline{SN}_-^{t\beta}(\emptyset)=\emptyset$, $\overline{SN}_-^{t\beta}(U)=U$.

(2) $\overline{SN}_+^{t\beta}(\emptyset)=\emptyset$, $\overline{SN}_+^{t\beta}(U)=U$.

(3) If $X\sqsubset Y$, then $\overline{SN}_-^{t\beta}(X)\sqsubset\overline{SN}_-^{t\beta}(Y)$ and $\overline{SN}_+^{t\beta}(X)\sqsubset\overline{SN}_+^{t\beta}(Y)$

(4) $\overline{SN}_-^{t\beta}(X)\sqcup\overline{SN}_-^{t\beta}(Y)\sqsubset\overline{SN}_-^{t\beta}(X\sqcup Y)$, $\overline{SN}_+^{t\beta}(X\sqcap Y)\sqsubset\overline{SN}_+^{t\beta}(X)\sqcap\overline{SN}_+^{t\beta}(Y)$.

(5) $\overline{SN}_-^{t\beta}(X)\sqcap\overline{SN}_-^{t\beta}(Y)=\overline{SN}_-^{t\beta}(X\sqcap Y)$, $\overline{SN}_+^{t\beta}(X\sqcup Y)=\overline{SN}_+^{t\beta}(X)\sqcup\overline{SN}_+^{t\beta}(Y)$.

(6) $\overline{SN}_-^{t\beta}(X^c)=(\overline{SN}_+^{t\beta}(X))^c$, $\overline{SN}_+^{t\beta}(X^c)=(\overline{SN}_-^{t\beta}(X))^c$.

(7) $\overline{SN}_-^{t\beta}(X)\sqsubset X\sqsubset\overline{SN}_+^{t\beta}(X)$.
\end{pro}

{\bf\slshape Proof} The proofs of (1)-(7) are similar to the proofs of Proposition \ref{pro411} (1)-(7).$\blacksquare$

\begin{pro} Let $\beta$ be a hesitant fuzzy number and $X,Y\sqsubset U$, the the following statements hold in $T_n$ for $x,y\in U$.

(1) $\overline{SN}_-^{n\beta}(\emptyset)=\emptyset$, $\overline{SN}_-^{n\beta}(U)=U$.

(2) $\overline{SN}_+^{n\beta}(\emptyset)=\emptyset$, $\overline{SN}_+^{n\beta}(U)=U$.

(3) If $X\sqsubset Y$, then $\overline{SN}_-^{n\beta}(X)\sqsubset\overline{SN}_-^{n\beta}(Y)$ and $\overline{SN}_+^{n\beta}(X)\sqsubset\overline{SN}_+^{n\beta}(Y)$

(4) $\overline{SN}_-^{n\beta}(X)\sqcup\overline{SN}_-^{n\beta}(Y)\sqsubset\overline{SN}_-^{n\beta}(X\sqcup Y)$, $\overline{SN}_+^{n\beta}(X\sqcap Y)\sqsubset\overline{SN}_+^{n\beta}(X)\sqcap\overline{SN}_+^{n\beta}(Y)$.

(5) $\overline{SN}_-^{n\beta}(X)\sqcap\overline{SN}_-^{n\beta}(Y)=\overline{SN}_-^{n\beta}(X\sqcap Y)$, $\overline{SN}_+^{n\beta}(X\sqcup Y)=\overline{SN}_+^{n\beta}(X)\sqcup\overline{SN}_+^{n\beta}(Y)$.

(6) $\overline{SN}_-^{n\beta}(X^c)=(\overline{SN}_+^{n\beta}(X))^c$, $\overline{SN}_+^{n\beta}(X^c)=(\overline{SN}_-^{n\beta}(X))^c$.

(7) $\overline{SN}_-^{n\beta}(X)\sqsubset X\sqsubset\overline{SN}_+^{n\beta}(X)$.

{\bf\slshape Proof} The proofs of (1)-(7) are similar to the proofs of Proposition \ref{pro411} (1)-(7).$\blacksquare$
\end{pro}

$x\in\overline{SN}^{m\beta}_x$ and $x\in\overline{SN}^{s\beta}_x$ are not necessarily true, then $\overline{SN}_-^{m\beta}(X)\sqsubset X\sqsubset\overline{SN}_+^{m\beta}(X)$ and $\overline{SN}_-^{s\beta}(X)\sqsubset X\sqsubset\overline{SN}_+^{s\beta}(X)$ are not necessarily true (as shown in Example \ref{exam410} (3) and (4)).

\begin{thm} Let $\beta$ be a hesitant fuzzy number, $X\sqsubset U$, $F:E\rightarrow HF(U)$, $A\sqsubset E$, $B\sqsubset E$.

(1) In $T_{p_1}=(U,F,A)_{\beta}$ and $T_{p_2}=(U,F,B)_{\beta}$, if $\overline{SN}^{p_1\beta}_x=\overline{SN}^{p_2\beta}_x$ for all $x\in U$, then $\overline{SN}_-^{p_1\beta}(X)=\overline{SN}_-^{p_2\beta}(X)$ and $\overline{SN}_+^{p_1\beta}(X)=\overline{SN}_+^{p_2\beta}(X)$.

(2) In $T_{a_1}=(U,F,A)_{\beta}$ and $T_{a_2}=(U,F,B)_{\beta}$, if $\overline{SN}^{a_1\beta}_x=\overline{SN}^{a_2\beta}_x$ for all $x\in U$, then $\overline{SN}_-^{a_1\beta}(X)=\overline{SN}_-^{a_2\beta}(X)$ and $\overline{SN}_+^{a_1\beta}(X)=\overline{SN}_+^{a_2\beta}(X)$.

(3) In $T_{m_1}=(U,F,A)_{\beta}$ and $T_{m_2}=(U,F,B)_{\beta}$, if $\overline{SN}^{m_1\beta}_x=\overline{SN}^{m_2\beta}_x$ for all $x\in U$, then $\overline{SN}_-^{m_1\beta}(X)=\overline{SN}_-^{m_2\beta}(X)$ and $\overline{SN}_+^{m_1\beta}(X)=\overline{SN}_+^{m_2\beta}(X)$.

(4) In $T_{s_1}=(U,F,A)_{\beta}$ and $T_{s_2}=(U,F,B)_{\beta}$, if $\overline{SN}^{s_1\beta}_x=\overline{SN}^{s_2\beta}_x$ for all $x\in U$, then $\overline{SN}_-^{s_1\beta}(X)=\overline{SN}_-^{s_2\beta}(X)$ and $\overline{SN}_+^{s_1\beta}(X)=\overline{SN}_+^{s_2\beta}(X)$.

(5) In $T_{t_1}=(U,F,A)_{\beta}$ and $T_{t_2}=(U,F,B)_{\beta}$, if $\overline{SN}^{t_1\beta}_x=\overline{SN}^{t_2\beta}_x$ for all $x\in U$, then $\overline{SN}_-^{t_1\beta}(X)=\overline{SN}_-^{t_2\beta}(X)$ and $\overline{SN}_+^{t_1\beta}(X)=\overline{SN}_+^{t_2\beta}(X)$.

(6) In $T_{n_1}=(U,F,A)_{\beta}$ and $T_{n_2}=(U,F,B)_{\beta}$, if $\overline{SN}^{n_1\beta}_x=\overline{SN}^{n_2\beta}_x$ for all $x\in U$, then $\overline{SN}_-^{n_1\beta}(X)=\overline{SN}_-^{n_2\beta}(X)$ and $\overline{SN}_+^{n_1\beta}(X)=\overline{SN}_+^{n_2\beta}(X)$.

{\bf\slshape Proof}  (1) Since $\overline{SN}^{p_1\beta}_x=\overline{SN}^{p_2\beta}_x$ for all $x\in U$, then
$\overline{SN}^{p_1\beta}_x\sqsubset X$ if and only if $\overline{SN}^{p_2\beta}_x\sqsubset X$, then $\overline{SN}_-^{p_1\beta}(X)=\overline{SN}_-^{p_2\beta}(X)$.

$\overline{SN}^{p_1\beta}_x\sqcap X\neq\emptyset$ if and only if $\overline{SN}^{p_2\beta}_x\sqcap X\neq\emptyset$, then $\overline{SN}_+^{p_1\beta}(X)=\overline{SN}_+^{p_2\beta}(X)$.

The proofs of (2)-(6) are similar to the proof of (1).$\blacksquare$
\end{thm}

\section{Conclusion}

Based on the inclusion relationships of hesitant fuzzy sets presented by Lu and Xu et al. \cite{LSZXZS}, this article studies some propositions of hesitant soft sets, hesitant fuzzy soft  $\beta$-neighborhoods, hesitant soft $\beta$-neighborhoods and hesitant fuzzy soft $\beta$-covering approximation spaces.

\section*{Acknowledgments}

The authors thank the editors and the anonymous reviewers for
their helpful comments and suggestions that have led to this improved
version of the paper. The work was supported by the National Natural Science Foundations of China (No. 71904043).

\section*{References}

\bibliographystyle{plain}
\bibliography{reference}

\begin{thebibliography}{10}

\bibitem{AWNJE}
W.~Abuasaker, J.~Nguyen, and et~al.
\newblock Perceptual maps to aggregate assessments from different rating
  profiles: A hesitant fuzzy linguistic approach.
\newblock {\em Applied Soft Computing}, 147:110803, 2023.

\bibitem{MANH}
M.~Akram, H.~S. Nawaz, and C.~Kahraman.
\newblock Rough pythagorean fuzzy approximations with neighborhood systems and
  information granulation.
\newblock {\em Expert Systems with Applications}, 218:119603, 2023.

\bibitem{JCVT}
J.~C.~R. Alcantud and V.~Torra.
\newblock Decomposition theorems and extension principles for hesitant fuzzy
  sets.
\newblock {\em Information Fusion}, 41:48--56, 2018.

\bibitem{BJ}
K.~V. Babitha and S.~J. John.
\newblock Hesitant fuzzy soft sets.
\newblock {\em Journal of New Results in Science}, 3:98--107, 2013.

\bibitem{BJAT}
J.~B{\l}aszczy{\'n}ski, A.~T. de~Almeida~Filho, and et~al.
\newblock Auto loan fraud detection using dominance-based rough set approach
  versus machine learning methods.
\newblock {\em Expert Systems with Applications}, 163:113740, 2021.

\bibitem{CHLT}
H.~Chen, T.~Li, and et~al.
\newblock A decision-theoretic rough set approach for dynamic data mining.
\newblock {\em IEEE Transactions on fuzzy Systems}, 23(6):1958--1970, 2015.

\bibitem{DYM}
H.~Din{\c{c}}er, S.~Y{\"{u}}ksel, and et~al.
\newblock Balanced scorecard-based analysis about european energy investment
  policies: A hybrid hesitant fuzzy decision-making approach with quality
  function deployment.
\newblock {\em Expert Systems With Applications}, 115:152--171, 2019.

\bibitem{JLGL}
J.~L. Garc{\'{\i}}a-Lapresta and D.~P{\'{e}}rez-Rom{\'{a}}n.
\newblock Consensus-based clustering under hesitant qualitative assessments.
\newblock {\em Fuzzy Sets and Systems}, 292:261--273, 2016.

\bibitem{GSKGA}
S.~K. Ghosh, A.~Ghosh, and et~al.
\newblock Recognition of cancer mediating biomarkers using rough approximations
  enabled intuitionistic fuzzy soft sets based similarity measure.
\newblock {\em Applied Soft Computing}, 124:109052, 2022.

\bibitem{HXZS}
Z.~Hao, Z.~Xu, and et~al.
\newblock Probabilistic dual hesitant fuzzy set and its application in risk
  evaluation.
\newblock {\em Knowledge-Based Systems}, 127:16--28, 2017.

\bibitem{HBGC}
B.~Huang, C.~Guo, and et~al.
\newblock Hierarchical structures and uncertainty measures for intuitionistic
  fuzzy approximation space.
\newblock {\em Information Sciences}, 336:92--114, 2016.

\bibitem{HBLIH}
B.~Huang, H.~Li, and et~al.
\newblock Intuitionistic fuzzy $\beta$-covering-based rough sets.
\newblock {\em Artificial Intelligence Review}, 53(4):2841--2873, 2020.

\bibitem{HJH}
H.~Jiang and B.~Q. Hu.
\newblock A decision-theoretic fuzzy rough set in hesitant fuzzy information
  systems and its application in multi-attribute decision-making.
\newblock {\em Information Sciences}, 579:103--127, 2021.

\bibitem{KES}
E.~Kannan, S.~Ravikumar, and et~al.
\newblock Analyzing uncertainty in cardiotocogram data for the prediction of
  fetal risks based on machine learning techniques using rough set.
\newblock {\em Journal of Ambient Intelligence and Humanized Computing}, pages
  1--13, 2021.

\bibitem{BLYY}
B.~Li, Y.~Yang, and et~al.
\newblock Two-sided matching model for complex product manufacturing tasks
  based on dual hesitant fuzzy preference information.
\newblock {\em Knowledge-Based Systems}, 186:104989, 2019.

\bibitem{HLJW}
H.~Li, J.~Wang, and et~al.
\newblock Air quality deterministic and probabilistic forecasting system based
  on hesitant fuzzy sets and nonlinear robust outlier correction.
\newblock {\em Knowledge-Based Systems}, 237:107789, 2022.

\bibitem{LLYX}
L.~Li and Y.~Xu.
\newblock An extended hesitant fuzzy set for modeling multi-source uncertainty
  and its applications in multiple-attribute decision-making.
\newblock {\em Expert Systems With Applications}, 238:121834, 2024.

\bibitem{LSZXZS}
S.~Lu, Z.~Xu, and et~al.
\newblock Foundational theories of hesitant fuzzy sets and hesitant fuzzy
  information systems and their applications for multi-strength intelligent
  classifiers.
\newblock {\em arXiv preprint arXiv:2311.04256}, 2023.

\bibitem{LNZQY}
N.~Luo, Q.~Zhang, and et~al.
\newblock Three-way multi-attribute decision-making under the double hierarchy
  hesitant fuzzy linguistic information system.
\newblock {\em Applied Soft Computing}, 154:111315, 2024.

\bibitem{MAMA}
A.~Mukherjee and A.~Mukherjee.
\newblock Interval-valued intuitionistic fuzzy soft rough approximation
  operators and their applications in decision making problem.
\newblock {\em Annals of Data Science}, 9(3):611--625, 2022.

\bibitem{PMNME}
M.~Pant and N.~Mehra.
\newblock Strong $(\alpha,k)$-cut and computational-based segmentation based
  novel hesitant fuzzy time series forecasting model.
\newblock {\em Applied Soft Computing}, 153:111251, 2024.

\bibitem{PALA82}
Z.~Pawlak.
\newblock Rough sets.
\newblock {\em International Journal of Applied Mathematics and Computer
  Science}, 11:341--356, 1982.

\bibitem{MTNA}
M.~Tishya and A.~Anitha.
\newblock Precipitation prediction by integrating rough set on fuzzy
  approximation space with deep learning techniques.
\newblock {\em Applied Soft Computing}, 139:110253, 2023.

\bibitem{TV}
V.~Torra.
\newblock Hesitant fuzzy sets.
\newblock {\em International Journal of Intelligent Systems}, 25:529--539,
  2010.

\bibitem{WGLT}
G.~Wang, T.~Li, and et~al.
\newblock Double-local rough sets for efficient data mining.
\newblock {\em Information Sciences}, 571:475--498, 2021.

\bibitem{WJLX}
J.~Wang and X.~Li.
\newblock An overlap function-based three-way intelligent decision model under
  interval-valued fuzzy information systems.
\newblock {\em Expert Systems with Applications}, 238:122036, 2024.

\bibitem{XX}
M.~Xia and Z.~Xu.
\newblock Hesitant fuzzy information aggregation in decision making.
\newblock {\em International Journal of Approximate Reasoning}, 52:395--407,
  2011.

\bibitem{GXLY}
G.~Xin and L.~Ying.
\newblock Multi-attribute decision-making based on comprehensive hesitant fuzzy
  entropy.
\newblock {\em Expert Systems With Applications}, 237:121459, 2024.

\bibitem{JYE}
J.~Ye, J.~Zhan, and Z.~Xu.
\newblock A novel multi-attribute decision-making method based on fuzzy rough
  sets.
\newblock {\em Computers \& Industrial Engineering}, 155:107136, 2021.

\bibitem{YUG}
G.~Yu.
\newblock Relationships between fuzzy approximation spaces and their
  uncertainty measures.
\newblock {\em Information Sciences}, 528:181--204, 2020.

\bibitem{ZHKDJ}
K.~Zhang and J.~Dai.
\newblock Redefined fuzzy rough set models in fuzzy $\beta$-covering group
  approximation spaces.
\newblock {\em Fuzzy Sets and Systems}, 442:109--154, 2022.

\bibitem{ZKDJ}
K.~Zhang and J.~Dai.
\newblock Three-way multi-criteria group decision-making method in a fuzzy
  $\beta$-covering group approximation space.
\newblock {\em Information Sciences}, 599:1--24, 2022.

\bibitem{ZKZJ}
K.~Zhang, J.~Zhan, and et~al.
\newblock On multicriteria decision-making method based on a fuzzy rough set
  model with fuzzy $\alpha$-neighborhoods.
\newblock {\em IEEE Transactions on Fuzzy Systems}, 29(9):2491--2505, 2020.

\bibitem{ZXHWJQ}
X.~Zhang and J.~Wang.
\newblock Fuzzy $\beta$-covering approximation spaces.
\newblock {\em International Journal of Approximate Reasoning}, 126:27–47,
  2020.

\bibitem{ZHLW}
H.~Zhou, W.~Li, and et~al.
\newblock Dynamic maintenance of updating rough approximations in
  interval-valued ordered decision systems.
\newblock {\em Applied Intelligence}, pages 1--18, 2023.

\end{thebibliography}







\end{multicols}
\end{document}